\documentclass{l4dc2025}

\title[Attention-enhanced Autoencoders for Weather Prediction]{Attention-Enhanced Convolutional Autoencoder and Structured Delay Embeddings for Weather Prediction}
\usepackage{comment}
\usepackage{times}
\usepackage{booktabs}
\usepackage{algorithm}
\usepackage{algorithmic}
\usepackage{amsmath}
\usepackage{amsfonts}
\usepackage{fancyhdr}

\coltauthor{\Name{Amirpasha Hedayat} \Email{ahedayat@umich.edu}\and
 \Name{Karthik Duraisamy} \Email{kdur@umich.edu}\\
 \addr  Department of Aerospace Engineering, University of Michigan, Ann Arbor, MI, USA.}

\jmlrvolume{}
\jmlryear{2025}
\jmlrproceedings{}{Preprint.}

\begin{document}

\maketitle

\pagestyle{fancy}        % use fancy headers/footers
\fancyhf{}               % clear all header and footer fields

\fancyhead[LE]{Hedayat \& Duraisamy}
\fancyhead[RO]{Attention-Enhanced Autoencoders for Weather Prediction}

% page number in the center of the footer
\fancyfoot[C]{\thepage}

% header rule
\renewcommand{\headrulewidth}{0.4pt}  % line under header

\vspace{-1.25cm}
\begin{abstract}%
Weather prediction is a quintessential problem involving the forecasting of a complex, nonlinear, and chaotic high-dimensional dynamical system. This work introduces an efficient reduced-order modeling (ROM) framework for short-range weather prediction and investigates fundamental questions in dimensionality reduction and reduced order modeling of such systems. Unlike recent AI-driven models, which require extensive computational resources, our framework prioritizes efficiency while achieving reasonable accuracy. Specifically, a ResNet-based convolutional autoencoder augmented by block attention modules is developed to reduce the dimensionality of high-dimensional weather data. Subsequently, a linear operator is learned in the time-delayed embedding of the latent space to efficiently capture the dynamics. Using the ERA5 reanalysis dataset, we demonstrate that this framework performs well in-distribution as evidenced by  effectively predicting weather patterns within training data periods. We also identify important limitations in generalizing to future states, particularly in maintaining prediction accuracy beyond the training window. Our analysis reveals that weather systems exhibit strong temporal correlations that can be effectively captured through linear operations in an appropriately constructed embedding space, and that projection error rather than inference error is the main bottleneck. These findings shed light on some key challenges in reduced-order modeling of chaotic systems and point toward opportunities for hybrid approaches that combine efficient reduced-order models as baselines with more sophisticated AI architectures, particularly for applications in long-term climate modeling where computational efficiency is paramount.
\end{abstract}

\vspace{-0.25cm}
\section{Introduction}

Weather and climate prediction has long been recognized as a grand challenge in science and engineering. Atmospheric dynamics is governed by complex, chaotic, nonlinear partial differential equations (PDEs) involving coupled processes across vast spatial and temporal scales. Accurate forecasting necessitates high-resolution models that capture fine-scale phenomena, leading to significant computational costs in traditional numerical weather prediction (NWP) methods \citep{Bauer2015}.
%In recent years, machine learning (ML) and artificial intelligence (AI) have emerged as powerful tools for modeling complex systems. 
Recently,  AI models for weather prediction, such as FourCastNet \citep{pathak2022fourcastnet}, GraphCast \citep{lam2022graphcast}, and ClimaX \citep{nguyen2023climax}, have demonstrated good predictive capabilities by leveraging deep neural networks within  architectures such as transformers and neural operators to learn spatiotemporal patterns from large datasets. However, these models require extensive computational resources for development, often involving thousands of GPU hours and large-scale parallel computing infrastructures.

Reduced Order Modeling (ROM) techniques aim to simplify high-dimensional systems by projecting them onto a lower-dimensional manifold while preserving essential dynamics. By combining attention-enhanced Convolutional Autoencoders (CAEs) for dimensionality reduction with simple time delayed linear models for evolution, we develop a framework that captures the general patterns and dynamics of weather systems with significantly reduced computational costs.
Motivated by the need for computationally efficient yet reasonably accurate forecasting methods, this paper takes a different approach, ROMs, to investigate fundamental questions about weather prediction and dimensionality reduction. While previous studies have focused primarily on improving prediction accuracy, we aim to understand: a) What are the key limiting factors in compressed representations of atmospheric states?
b) How effectively can temporal correlations in weather patterns be captured through linear operations in appropriate embedding spaces?
c) What role does the quality of dimensional reduction play versus the sophistication of the prediction mechanism?
d) How well do traditional error metrics align with meteorologically significant features?

Our approach is also motivated by the broader need to develop computationally efficient baseline models that can complement existing state-of-the-art AI systems. By establishing and analyzing such baselines, we can better understand the potential for hybrid approaches that combine the efficiency of reduced-order models with the accuracy of more complex AI architectures. This understanding becomes particularly crucial when considering applications to long-term climate modeling, where computational efficiency remains a critical constraint.

 %The key contributions of this work are:
%\begin{itemize}
%    \item \textbf{Attention-based CAE}: Development of a ResNet-based CAE architecture enhanced with Convolutional Block Attention Modules (CBAM) to effectively compress high-dimensional weather data into a compact latent space.
 %   \item \textbf{Time-delayed Structured Embeddings}: Implementation of a time-delay-enhanced linear model inferred in a reduced-dimension latent space that efficiently learns the dynamics within the training data.
 %   \item \textbf{Evaluation of Generalization Capabilities}: Comprehensive experiments assessing the model's ability to interpolate within the training data and its limitations in generalizing to unseen future states, highlighting the importance of the latent representation.
%\end{itemize}

The remainder of the paper is organized as follows: Section \ref{sec:related_work} reviews related work in AI-based weather prediction and reduced-order modeling. Section \ref{sec:methodology} describes the proposed methodology, including detailed formulations of CBAM and the time-delayed linear model. Section \ref{sec:results} presents experimental results and analyses, discussing the limitations and key insights. Section \ref{sec:conclusion} concludes the paper with the main takeaways and future research directions.
\vspace{-0.25cm}
\section{Related Work}
\label{sec:related_work}

The integration of machine learning (ML) and artificial intelligence (AI) into weather prediction has led to significant advancements in forecasting accuracy and computational efficiency \citep{reichstein2019deep}. One of the notable AI-based weather models is FourCastNet, which employs a Fourier-based neural network architecture to efficiently capture global weather patterns by leveraging spectral convolutions adept at handling the periodic nature of atmospheric data \citep{pathak2022fourcastnet}. GraphCast utilizes graph neural networks (GNNs) to model spatial dependencies inherent in weather systems, effectively capturing localized interactions and long-range dependencies \citep{lam2022graphcast}. ClimaX and its newer version, Aurora, adopt a Transformer-based architecture to build a weather foundation model, excelling in modeling long-term dependencies and complex spatiotemporal relationships within weather data \citep{nguyen2023climax, aurora}. Despite their impressive performance, these AI-driven models often require substantial computational resources. This high computational demand poses challenges for scalability and real-time application, especially in resource-constrained environments.

Reduced-order modeling (ROM) aims to simplify high-dimensional dynamical systems by projecting them onto a lower-dimensional manifold while preserving essential dynamical characteristics. Traditional intrusive, projection-based ROM techniques, such as Galerkin and Petrov-Galerkin methods, have been widely used in various fields, including fluid dynamics and structural engineering \citep{Carlberg2011, Benner2015, HUANG2022110742}. %However, these methods require access to the full-order model, which is rarely the case in applications such as weather prediction. 
Data-driven approaches including Operator Inference \citep{Peherstorfer2016, pan2018data, BENNER2020113433, GEELEN2023115717, Kramer2024} have gained prominence for their ability to build efficient models in the absence of full-order models. %Operator inference technique formulates the reduced-order model by inferring the underlying operators that govern the system's dynamics directly from data, making it an ideal model for tasks like weather prediction where plenty of data exist.

Convolutional Autoencoders (CAEs) have been employed for dimensionality reduction in high-dimensional datasets \citep{hinton2006reducing, masci2011stacked, Mao2016, Theis2017,xu2020multi,pan2020physics}. By leveraging convolutional layers, CAEs can effectively capture spatial hierarchies and local dependencies, making them suitable for applications involving spatially structured data such as weather fields. These structures can be further improved by adding attention layers in the network. Attention mechanisms have revolutionized the field of deep learning by enabling models to focus on the most relevant parts of the input data, improving performance in various tasks. The concept was popularized in natural language processing (NLP) with the introduction of the Transformer architecture \citep{vaswani2017attention}, which relies entirely on self-attention mechanisms to capture dependencies between input elements. In computer vision, attention mechanisms have been incorporated to enhance feature representation in convolutional neural networks (CNNs). The Squeeze-and-Excitation (SE) network \citep{hu2018squeeze} introduced channel-wise attention by adaptively recalibrating channel-wise feature responses. Building upon this, the Convolutional Block Attention Module (CBAM) \citep{woo2018cbam} sequentially applies both channel and spatial attention, allowing the network to focus on \emph{what} and \emph{where} to emphasize or suppress.

Our work builds upon these foundations by developing a ResNet-based \citep{he2016deep} CAE enhanced with CBAM to achieve effective dimensionality reduction of high-dimensional weather data. By inferring a time-delay-enhanced linear model from the encoded space, we aim to evaluate the model's ability to learn the dynamics within the training data and assess its generalization capabilities to unseen future states.
\vspace{-0.25cm}
\section{Methodology}
\label{sec:methodology}

This section outlines the proposed ROM framework for weather prediction. The framework comprises two main components: a CAE with local attention mechanism for dimensionality reduction, and a time-delayed linear model for capturing latent-space dynamics.
\vspace{-0.25cm}
\subsection{Dataset}

We utilize the ERA5 dataset \citep{hersbach2020era5}, a comprehensive climate reanalysis dataset encompassing various atmospheric variables over an extensive temporal span. For this study, we focus on four key variables: U-component of wind at 10 meters: $u_{10}$;
        V-component of wind at 10 meters: $v_{10}$;
        Temperature at 2 meters: $T_{2m}$;
        Mean sea level pressure: $P_{msl}$.
% as shown in Table \ref{tab:dataset_variables}.
\begin{comment}

\begin{table}[ht]
    \centering
    \caption{Selected ERA5 Variables for Weather Prediction}
    \label{tab:dataset_variables}
    \begin{tabular}{lc}
        \toprule
        \textbf{Variable} & \textbf{Symbol} \\
        \midrule
        U-component of wind at 10 meters & $u_{10}$ \\
        V-component of wind at 10 meters & $v_{10}$ \\
        Temperature at 2 meters & $T_{2m}$ \\
        Mean sea level pressure & $P_{msl}$ \\
        \bottomrule
    \end{tabular}
\end{table}
\end{comment}
The data is spatially resolved at a resolution of $240 \times 121$ (longitude $\times$ latitude) and temporally sampled every six hours from 2011 to 2022. Each variable is normalized to have zero mean and unit variance to facilitate stable training. The dataset is split into training and test sets, with data from 12/1/2011 to 12/1/2021 used for training and data from 12/1/2021 to 12/31/2021 reserved for testing.
\vspace{-0.25cm}
\subsection{Convolutional Autoencoder with Attention Mechanism}

To address the high dimensionality of the ERA5 dataset, we employ a CAE enhanced with CBAM \citep{woo2018cbam}. The CAE compresses the input data into a lower-dimensional latent space while preserving essential dynamical information, and CBAM, as a lightweight attention module, sequentially applies channel and spatial attention to refine feature maps, enhancing the network's representational power. Given an intermediate feature map $\mathbf{F} \in \mathbb{R}^{C \times H \times W}$, CBAM computes attention maps along the channel and spatial dimensions and refines $\mathbf{F}$ accordingly, making it context-aware.

\noindent\textbf{Channel Attention Module:} Emphasizes informative channels by modeling inter-channel relationships. It computes a channel attention map $\mathbf{M}_{\text{c}} \in \mathbb{R}^{C \times 1 \times 1}$ using global average pooling and max pooling followed by a shared multi-layer perceptron (MLP). The refined feature map is obtained by:
\begin{align}
\mathbf{M}_{\text{c}} &= \sigma\left(\text{MLP}(\text{AvgPool}_{\text{global}}(\mathbf{F})) + \text{MLP}(\text{MaxPool}_{\text{global}}(\mathbf{F}))\right), \ \ 
\mathbf{F}' = \mathbf{M}_{\text{c}} \otimes \mathbf{F},
\end{align}
where $\sigma$ denotes the sigmoid function and $\otimes$ denotes element-wise multiplication with broadcasting.

\noindent\textbf{Spatial Attention Module:} Focuses on important spatial regions by modeling inter-spatial relationships. It computes a spatial attention map $\mathbf{M}_{\text{s}} \in \mathbb{R}^{1 \times H \times W}$ using average and max pooling across the channel dimension, followed by a convolutional layer:
\begin{align}
\mathbf{M}_{\text{s}} &= \sigma\left(f^{7 \times 7}\left([\text{AvgPool}_{\text{channel}}(\mathbf{F}'); \text{MaxPool}_{\text{channel}}(\mathbf{F}')]\right)\right), \ \
\mathbf{F}'' = \mathbf{M}_{\text{s}} \otimes \mathbf{F}'.
\end{align}

\begin{figure}[ht]
    \centering
    \includegraphics[width=0.8\linewidth]{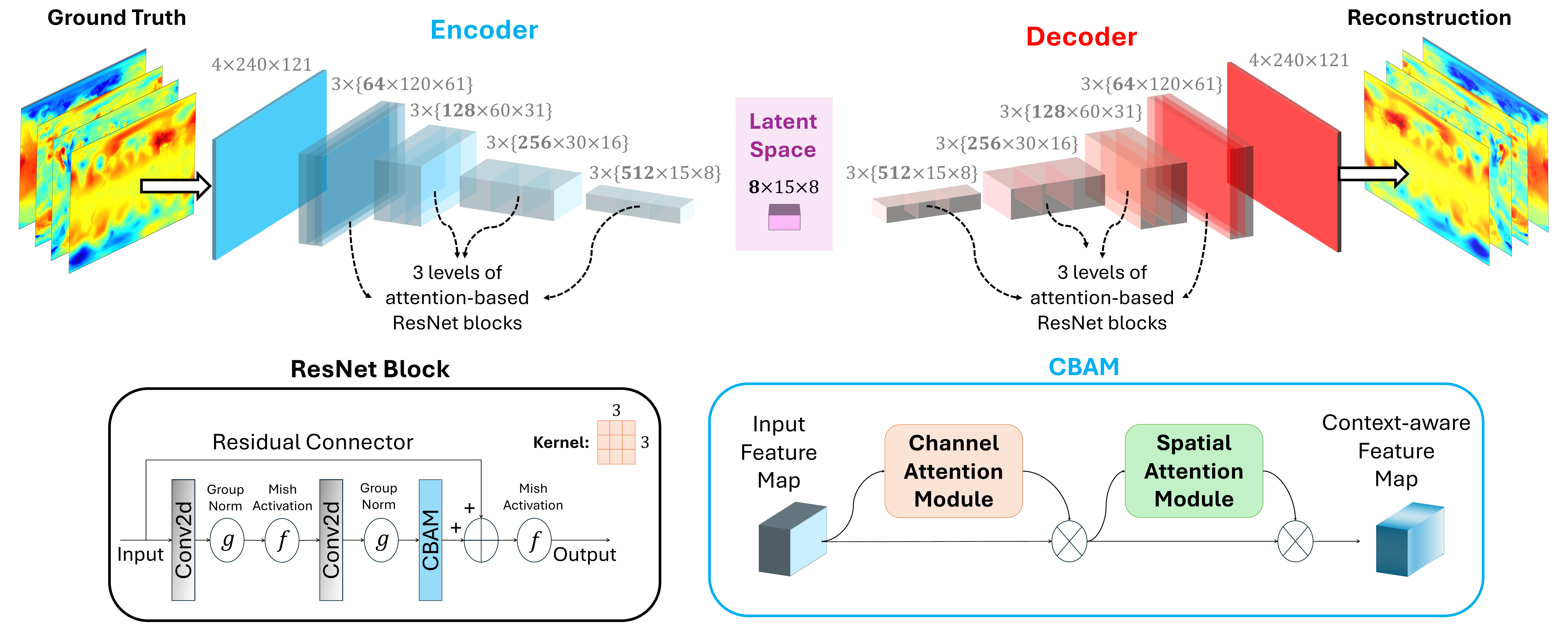}
    \vspace{-0.25cm}
    \caption{Architecture of the CBAM-enhanced CAE, having 31.72M trainable parameters.}
    \label{fig:cae_architecture}
\end{figure}

Given an input tensor $\mathbf{X} \in \mathbb{R}^{C \times H \times W}$, where $C$ is the number of variables (channels), and $H$ and $W$ are the spatial dimensions, the CAE aims to learn an encoding function $\mathcal{E}: \mathbb{R}^{C \times H \times W} \rightarrow \mathbb{R}^{c \times h \times w}$ and a decoding function $\mathcal{D}: \mathbb{R}^{c \times h \times w} \rightarrow \mathbb{R}^{C \times H \times W}$. The CAE is trained to minimize the discrepancy between the input, $\mathbf{X}$, and its reconstruction, $\hat{\mathbf{X}} = \mathcal{D}(\mathcal{E}(\mathbf{X}))$, using the latitude-weighted root mean square error (LW-RMSE) to account for the non-uniform distribution of grid points due to the spherical nature of Earth:
\begin{align}
\mathcal{L}_{\text{LW-RMSE}} = \sqrt{\frac{1}{N} \sum_{i=1}^{N} w(\phi_i) \left( \mathbf{X}_i - \hat{\mathbf{X}}_i \right)^2},
\end{align}
where $N$ is the total number of spatial grid points, $\phi_i$ is the latitude of grid point $i$, and $w(\phi_i) = \cos(\phi_i) / \left( \frac{1}{M} \sum_{j=1}^{M} \cos(\phi_j) \right)$ is the weighting factor with $M$ being the number of resolved latitude angles. Training is performed using the Adam optimizer with an initial learning rate of $1 \times 10^{-3}$, reduced on plateau, and a batch size of 32 for 100 epochs. The overall architecture of the developed CAE, as well as the components of the ResNet block and CBAM layer are presented in Figure \ref{fig:cae_architecture}.

\vspace{-0.5cm}
\subsection{Time-delayed Operators}

Consider a high-dimensional dynamical system characterized by state variables $\mathbf{x}(t) \in \mathbb{R}^{N_{\text{high}}}$. To efficiently model the temporal evolution of this system, the attention-based CAE model from the previous section can be used to get the sequence of latent states denoted by $\{\mathbf{z}_k\}_{k=1}^N$, where each $\mathbf{z}_k \in \mathbb{R}^{n}$ with $n \ll N_{\text{high}}$. In the standard operator inference framework, as developed in \citep{Peherstorfer2016}, the continuous-time evolution of the latent state $\mathbf{z}(t)$ is modeled by an ordinary differential equation (ODE):
\vspace{-0.25cm}
\begin{align}
\frac{d\mathbf{z}(t)}{dt} = \mathbf{A} \mathbf{z}(t) + \sum_{j=2}^{p} \mathbf{H}^{(j)} \left( \mathbf{z}(t)^{\otimes j} \right),
\end{align}
where $\mathbf{A} \in \mathbb{R}^{n \times n}$ represents the linear operator, $\mathbf{H}^{(j)} \in \mathbb{R}^{n \times n^j}$ are higher-order tensors capturing polynomial nonlinearities up to degree $p$, and $\mathbf{z}(t)^{\otimes j}$ denotes the $j$-fold Kronecker product of $\mathbf{z}(t)$ that accounts for the commutativity of the multiplications. The objective is to infer these operators directly from data without intrusive access to the underlying governing equations. However, estimating high-order tensors $\mathbf{H}^{(j)}$ becomes computationally intractable as $p$ increases, due to the exponential growth of the tensor dimensions. To alleviate this issue and to exploit  the availability of past time snapshots and correlations therein, we construct a time-delayed latent state vector $\mathbf{z}_k^{\text{td}} \in \mathbb{R}^{nd}$ defined as:
$
\mathbf{z}_k^{\text{td}} = \begin{bmatrix} \mathbf{z}_k^\top & \mathbf{z}_{k-1}^\top & \cdots & \mathbf{z}_{k-d+1}^\top \end{bmatrix}^\top,
$
where $d \in \mathbb{N}$ is the embedding dimension, representing the number of delay coordinates. This embedding reconstructs the state-space trajectory in a higher-dimensional space, allowing for the approximation of nonlinear dynamics via linear operators \citep{takens1981detecting,Pan2020}.
We propose a discrete-time linear model to predict the future state:
$
\mathbf{z}_{k+1} = \mathbf{L} \mathbf{z}_k^{\text{td}} + \boldsymbol{\varepsilon}_k,
$
where $\mathbf{L} \in \mathbb{R}^{n \times nd}$ is the linear operator to be inferred, and $\boldsymbol{\varepsilon}_k \in \mathbb{R}^n$ denotes the modeling error at time step $k$. The goal is to find $\mathbf{L}$ that minimizes the cumulative prediction error over the available data. Therefore, the operator $\mathbf{L}$ can be obtained by solving the following optimization problem: $\mathbf{L}^\ast = \underset{\mathbf{L} \in \mathbb{R}^{n \times nd}}{\operatorname{argmin}} \; \left\| \mathbf{Z}_{\text{future}} - \mathbf{L} \mathbf{Z}_{\text{td}} \right\|_{F}^{2}$, where  $\|\cdot\|_{F}$ denotes the Frobenius norm, and the data matrices are given by
$
\mathbf{Z}_{\text{future}} = \begin{bmatrix} \mathbf{z}_{d+1} & \mathbf{z}_{d+2} & \cdots & \mathbf{z}_N \end{bmatrix} \in \mathbb{R}^{n \times (N - d)}$ and $
\mathbf{Z}_{\text{td}} = \begin{bmatrix} \mathbf{z}_d^{\text{td}} & \mathbf{z}_{d+1}^{\text{td}} & \cdots & \mathbf{z}_{N-1}^{\text{td}} \end{bmatrix} \in \mathbb{R}^{nd \times (N - d)}.
$
 %Having trained the attention-enhanced CAE and the time-delayed model on ERA5 dataset, the prediction workflow is shown in Algorithm \ref{alg:prediction_workflow}.

\begin{comment}
\begin{algorithm}[H]
\caption{CAE-encoded ROM Prediction Workflow}
\label{alg:prediction_workflow}
\begin{algorithmic}[1]
\REQUIRE Normalized initial state $\mathbf{X}_0$ and $d-1$ delayed states $\{\mathbf{X}_{-i}\}_{i=1}^{d-1}$, trained CAE encoder $\mathcal{E}$, trained CAE decoder $\mathcal{D}$, inferred operator $\mathbf{L}^\ast$, and forecast horizon $T$.
\STATE Encode the initial and delayed states into the latent space:
$
\{\mathbf{z}_{-i}\}_{i=0}^{d-1} = \{\mathcal{E}(\mathbf{X}_{-i})\}_{i=0}^{d-1}.
$
\STATE Construct the initial time-delayed latent state:
$
\mathbf{z}_0^{\text{td}} = \begin{bmatrix} \mathbf{z}_0^\top & \mathbf{z}_{-1}^\top & \cdots & \mathbf{z}_{-d+1}^\top \end{bmatrix}^\top.
$
\FOR{$t = 0$ to $T-1$}
    \STATE Predict the next latent state using the inferred model:
    $
    \mathbf{z}_{t+1} = \mathbf{L}^\ast \mathbf{z}_t^{\text{td}}.
    $
    \STATE Update the time-delayed latent state:
    $
    \mathbf{z}_{t+1}^{\text{td}} = \begin{bmatrix} \mathbf{z}_{t+1}^\top & \mathbf{z}_t^\top & \cdots & \mathbf{z}_{t-d+2}^\top \end{bmatrix}^\top.
    $
\ENDFOR
\STATE Decode the predicted latent states back to the physical space:
$
\{\hat{\mathbf{X}}_t\}_{t=1}^{T} = \{\mathcal{D}(\mathbf{z}_t)\}_{t=1}^{T}.
$
\RETURN Predicted normalized physical states $\{\hat{\mathbf{X}}_t\}_{t=1}^{T}$.
\end{algorithmic}
\end{algorithm}
\vspace{-0.85cm}
\end{comment}
\vspace{-0.25cm}
\section{Results and Discussion}
\label{sec:results}

In this section, we present evaluations  of the proposed ROM framework. We first assess the dimensionality reduction capability of the CAE compared to Proper Orthogonal Decomposition (POD), highlighting the advantages of nonlinear dimensionality reduction. Then, we evaluate the predictive performance of the ROM through three key experiments, analyzing its ability to interpolate within the training data and its limitations in generalizing to unseen future states. Due to space limitations, we do not cover extensive hyper-parameter optimization and related explorations.
\vspace{-0.25cm}
\subsection{CAE Reconstruction Results}

\begin{figure}[ht]
    \centering
    \includegraphics[width=0.4\textwidth]{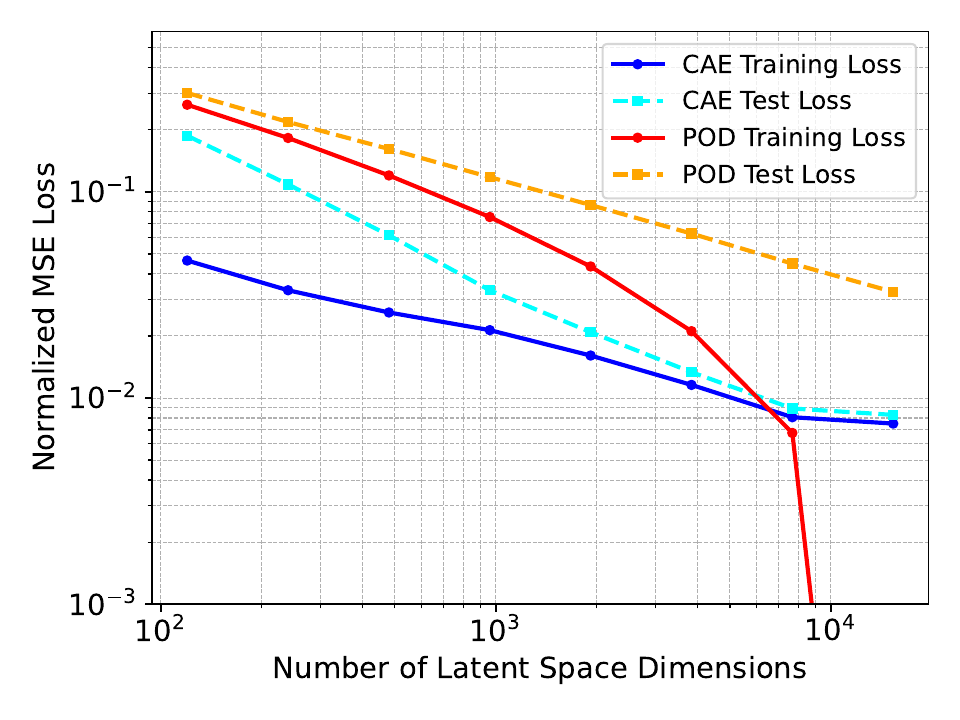}%
\vspace{-0.7cm}
\caption{Comparing dimensionality reduction on ERA5 as a function of the latent space dimension.}
    \label{fig:era5-red}
\end{figure}

Proper Orthogonal Decomposition (POD)~\citep{berkooz1993proper} is a linear dimensionality reduction technique widely used in reduced-order modeling. POD decomposes the data into orthogonal modes ranked by their energy content, and subsequently reduces the dimensionality by truncating modes with lower energy. We conducted an analysis to investigate the effect of the latent dimension on the loss magnitude, for both POD and CAE as shown in Fig. \ref{fig:era5-red}. Comparing CAE's performance against POD for the out-of-distribution samples, we observe that CAE results in less information loss. The slow decay indicates that the underlying dynamics of the weather dataset are highly complex and nonlinear, and a latent dimension on the order of $\mathcal{O}(10^4)$ is necessary for the loss to reach a plateau in the CAE's case. This requirement imposes substantial computational demands, undermining the efficiency goals in ROM development. To strike a balance between computational efficiency and representational accuracy, a latent space dimension of 960 (8 channels in the CAE’s latent space) was selected. This choice provides an acceptable compromise, enabling the CAE to capture the critical features of the dataset without incurring excessive computational cost.

\begin{figure}[ht]
    \centering
    \subfigure[POD out-of-distribution reconstruction]{
        \includegraphics[width=0.4\textwidth,height=\textheight,keepaspectratio]{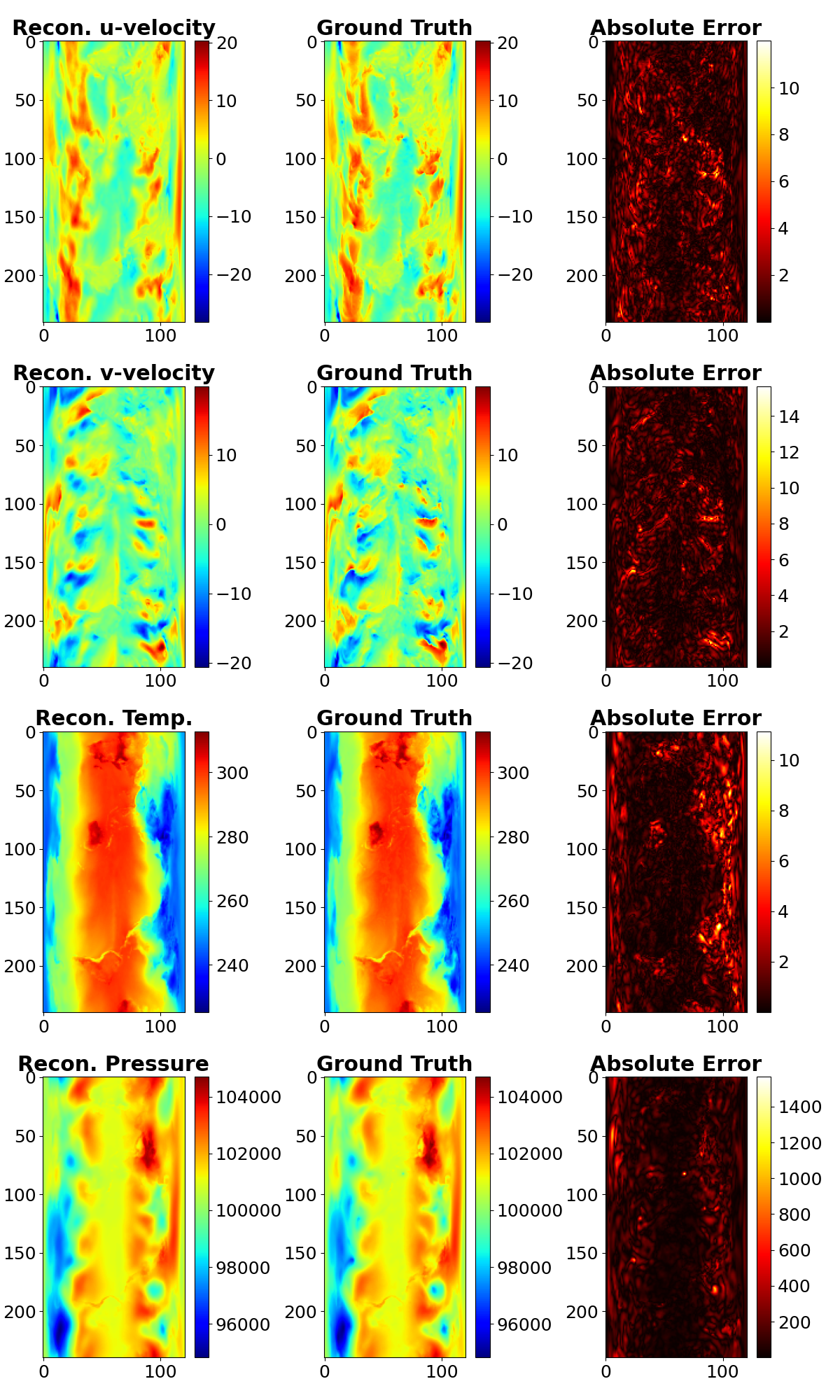}%
        \label{fig:pod_recon}}
    % \hfill
    \subfigure[CAE out-of-distribution reconstruction]{
        \includegraphics[width=0.4\textwidth,height=\textheight,keepaspectratio]{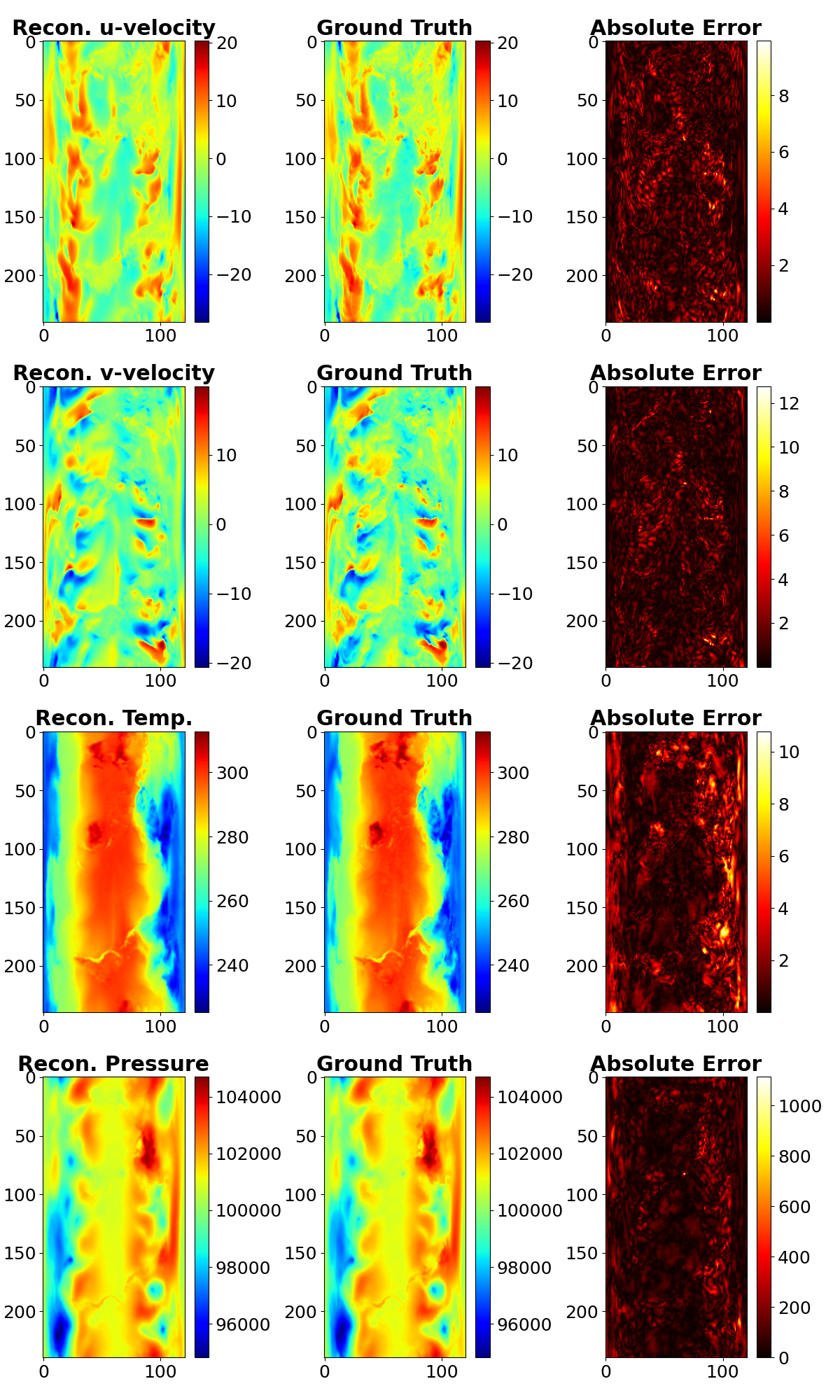}%
        \label{fig:cae_recon}}
    \subfigure[POD LW-RMSE]{
        \includegraphics[width=0.4\textwidth,height=\textheight,keepaspectratio]{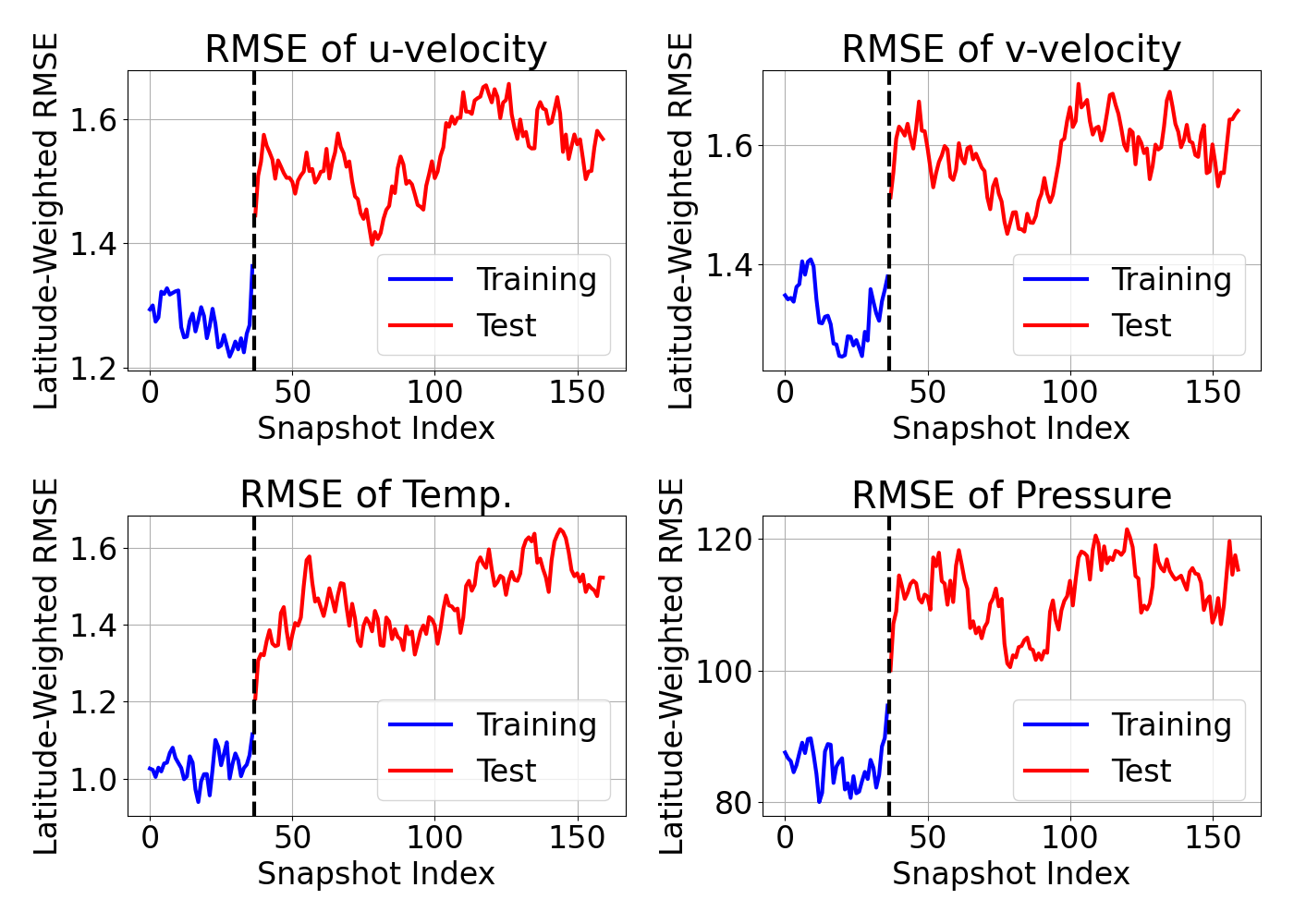}%
        \label{fig:pod_rmse}}
    % \hfill
    \subfigure[CAE LW-RMSE]{
        \includegraphics[width=0.4\textwidth,height=\textheight,keepaspectratio]{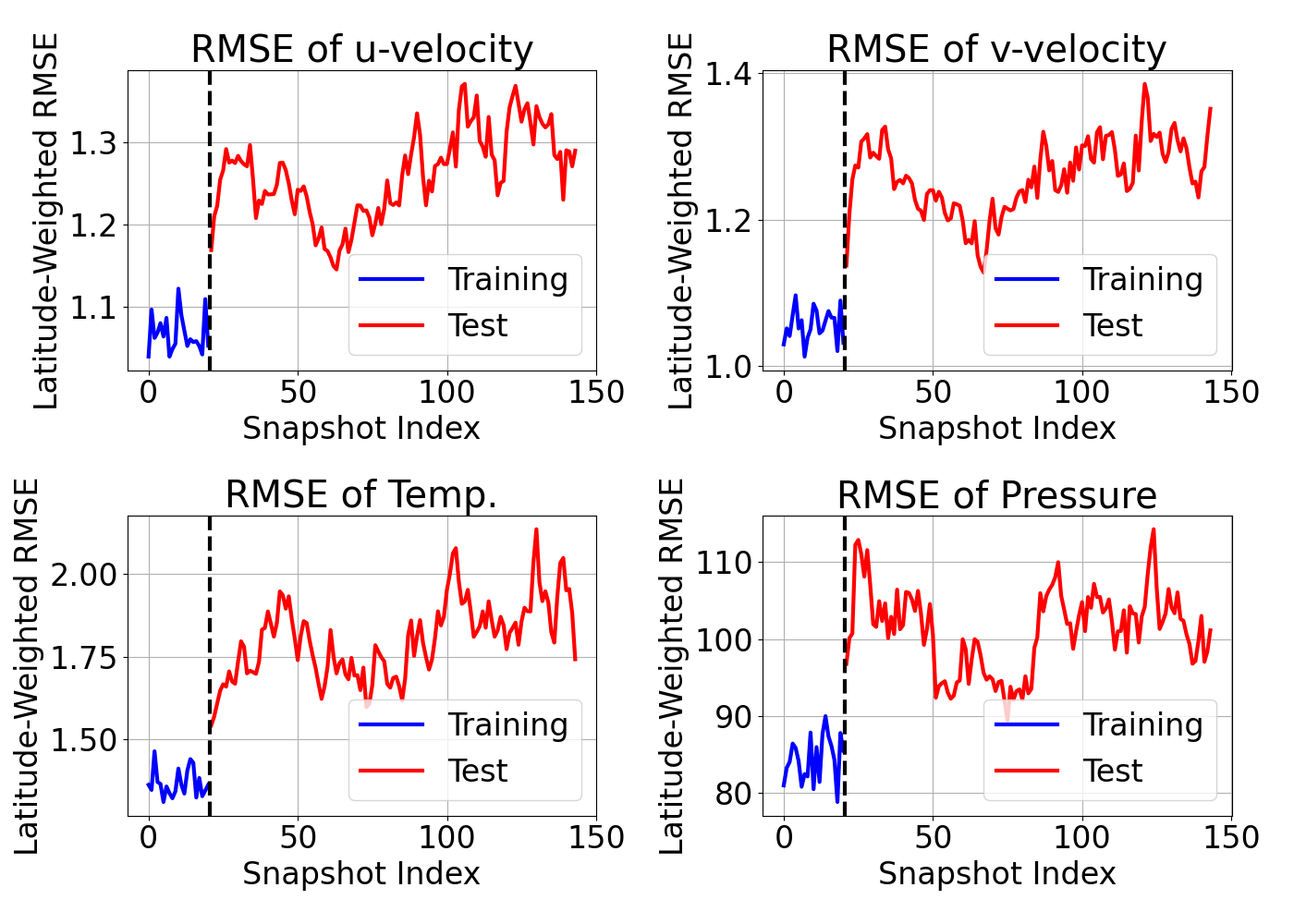}%
        \label{fig:cae_rmse}}
    \vspace{-0.25cm}
    \caption{Comparing ERA5 dimensionality reduction using CAE against POD.}
    \label{fig:dim-red}
\end{figure}

% \vspace{-0.7cm}
The CAE, enhanced with CBAM leverages nonlinear transformations and attention mechanisms to focus on relevant features, resulting in a slightly more efficient representation compared to  linear dimensionality reduction. We compare the out-of-distribution reconstruction performance of the CAE and POD using LW-RMSE. Both methods are configured to achieve a roughly similar compression ratio, and the performance of each in reconstructing the target atmospheric variables is tabulated in Table \ref{tab:reconstruction_rmse}.
\begin{table}[ht]
    \centering
    \vspace{-0.5cm}
    \caption{Average Test-set Reconstruction LW-RMSE Comparison between CAE and POD}
    \label{tab:reconstruction_rmse}
    \begin{tabular}{lccccc}
        \toprule
        \textbf{Model} & \textbf{Comp. Ratio} & $u_{10}(m/s)$ & $v_{10}(m/s)$ & $T_{2m}(K)$ & $P_{msl} (Pa)$ \\
        \midrule
        POD (1000 modes) & 121:1 & 1.55 & 1.6 & \textbf{1.5} & 110 \\
        CAE (960 latent dimensions) & 121:1 & \textbf{1.25} & \textbf{1.25} & 1.9 & \textbf{102} \\
        \bottomrule
    \end{tabular}
\end{table}
The CAE achieves a lower reconstruction error compared to POD for all variables except the temperature, with similar compression ratio.  Figure \ref{fig:dim-red} visually compares the reconstruction of all four variables using POD and the CAE. The CAE reconstruction retains more fine-scale features and closely resembles the ground truth, while the POD reconstruction exhibits smoothing and loss of detail in some parts of the domain. The difference is more visible when inspecting wind components. These results justify the use of advanced nonlinear dimensionality reduction techniques such as the CAE in our ROM framework, providing a strong foundation for modeling the system's dynamics in the latent space.

Moreover, an ablation study was conducted to evaluate the impact of the CBAM layers on the network's performance. For this analysis, the attention layers in the ResNet blocks were deactivated to assess their effect on the reconstruction LW-RMSE of the samples. The results, presented in Fig. \ref{fig:cbam}, demonstrate that incorporating CBAM indeed enhances the model's reconstruction capability for both in-distribution and out-of-distribution samples across all atmospheric variables.

\begin{figure}[ht]
    \centering
    \subfigure[CAE with CBAM]{
        \includegraphics[width=0.4\textwidth,height=\textheight,keepaspectratio]{Images/cae/WIth_CBAM.png}%
        \label{fig:cbam}}
    % \hfill
    \subfigure[CAE without CBAM]{
        \includegraphics[width=0.4\textwidth,height=\textheight,keepaspectratio]{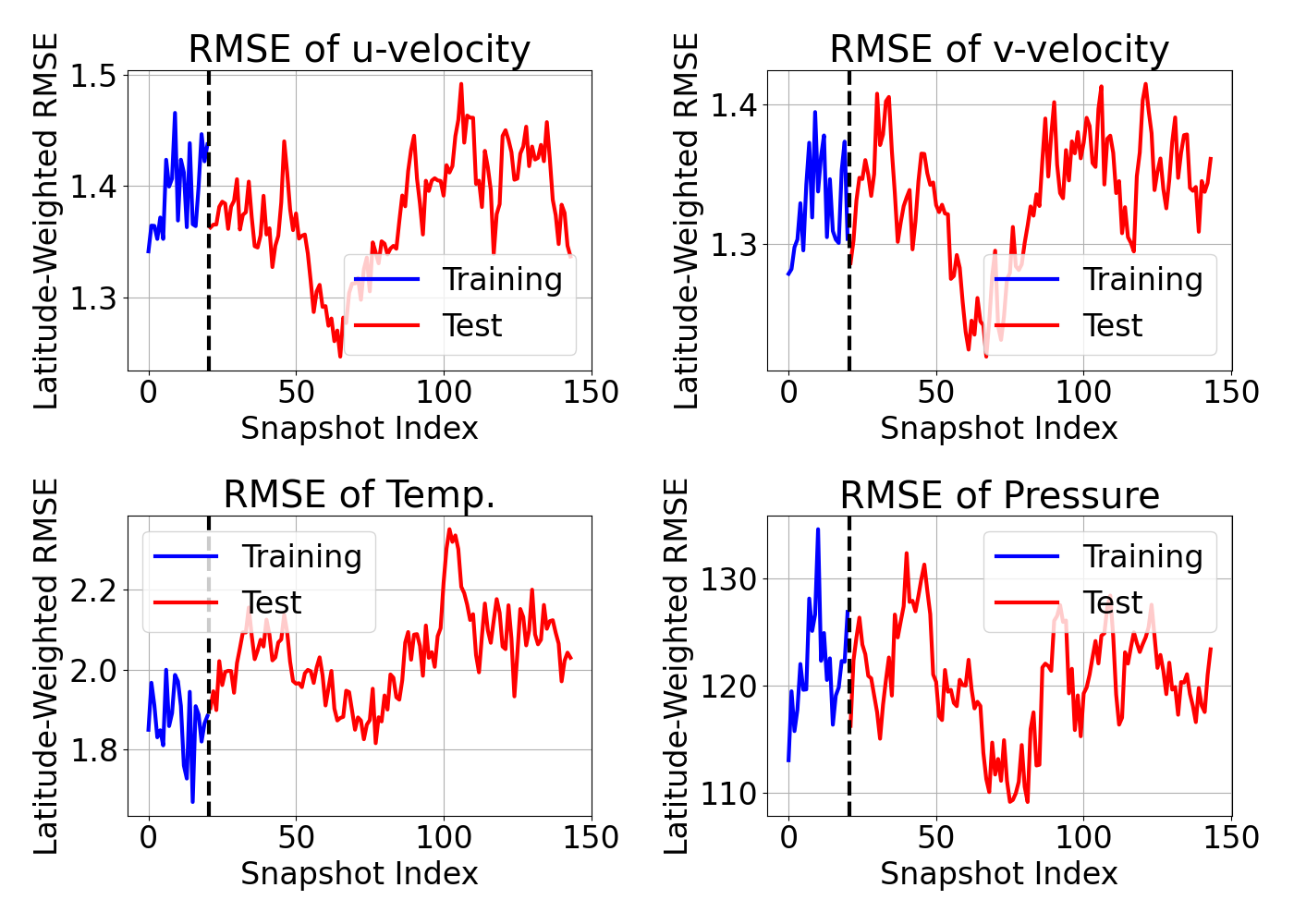}%
        \label{fig:cbam}}
    \vspace{-0.25cm}
    \caption{Effect of CBAM layers on the reconstruction LW-RMSE.}
    \label{fig:cbam}
\end{figure}
\vspace{-0.25cm}
\subsection{ROM Prediction Results}
\begin{figure}[ht]
    \centering
    \subfigure[u-Velocity prediction ($d=15$)]{
        \includegraphics[width=0.4\textwidth,height=\textheight,keepaspectratio]{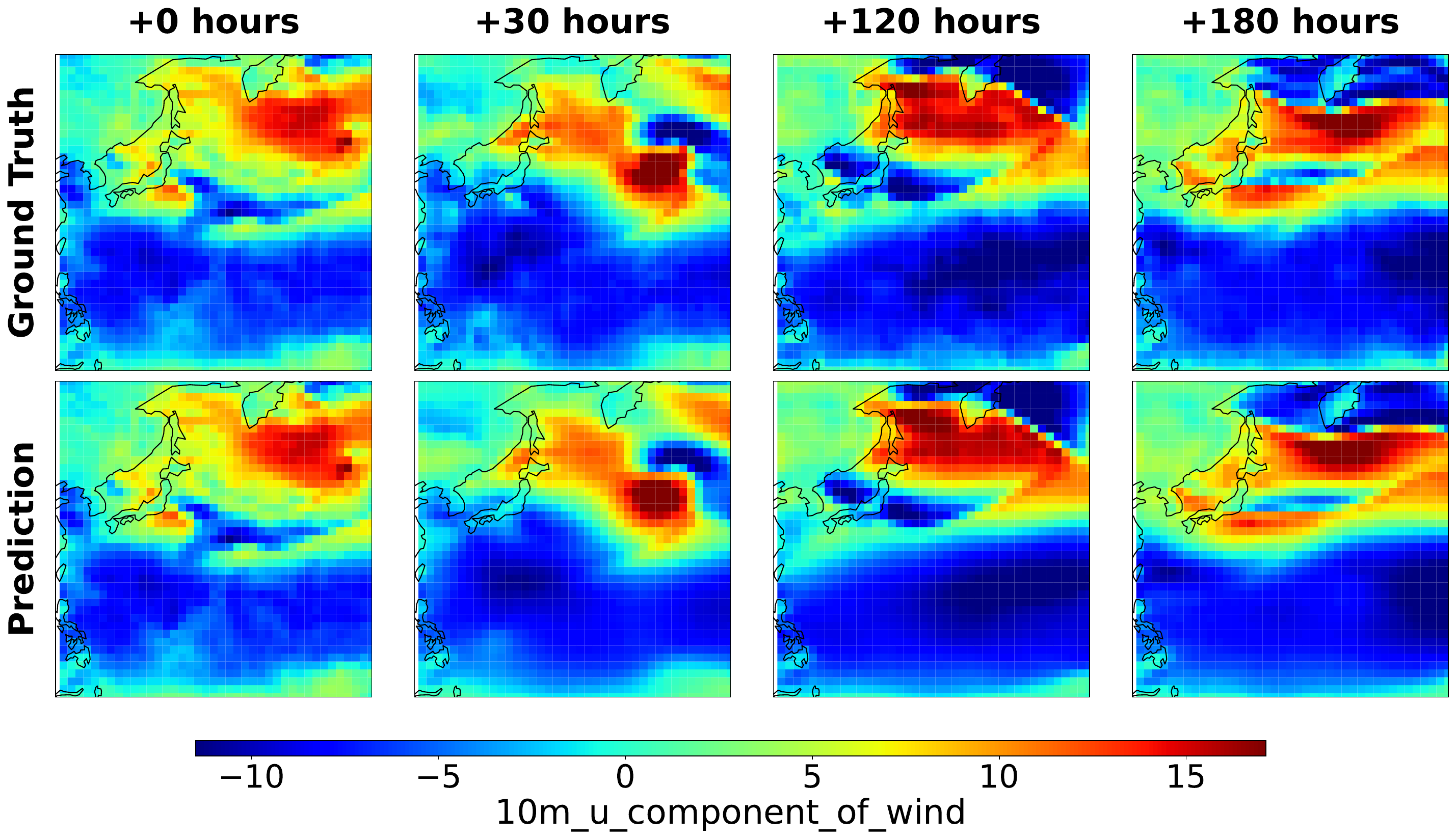}%
        \label{fig:opinf_exp1_u}}
    % \hfill
    \subfigure[v-Velocity prediction ($d=15$)]{
        \includegraphics[width=0.4\textwidth,height=\textheight,keepaspectratio]{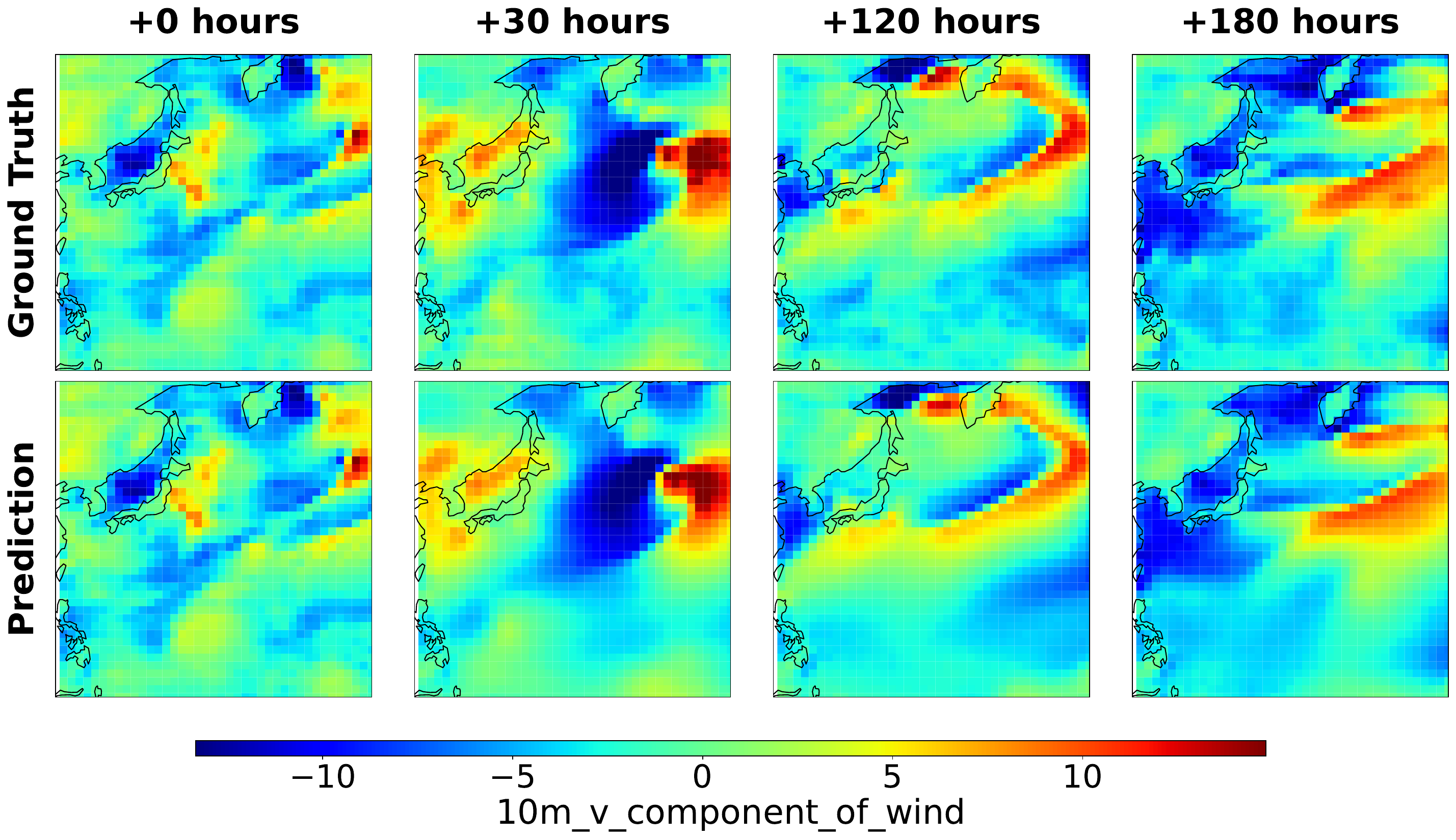}%
        \label{fig:opinf_exp1_v}}
    % \hfill
    \subfigure[Temperature prediction ($d=15$)]{
        \includegraphics[width=0.4\textwidth,height=\textheight,keepaspectratio]{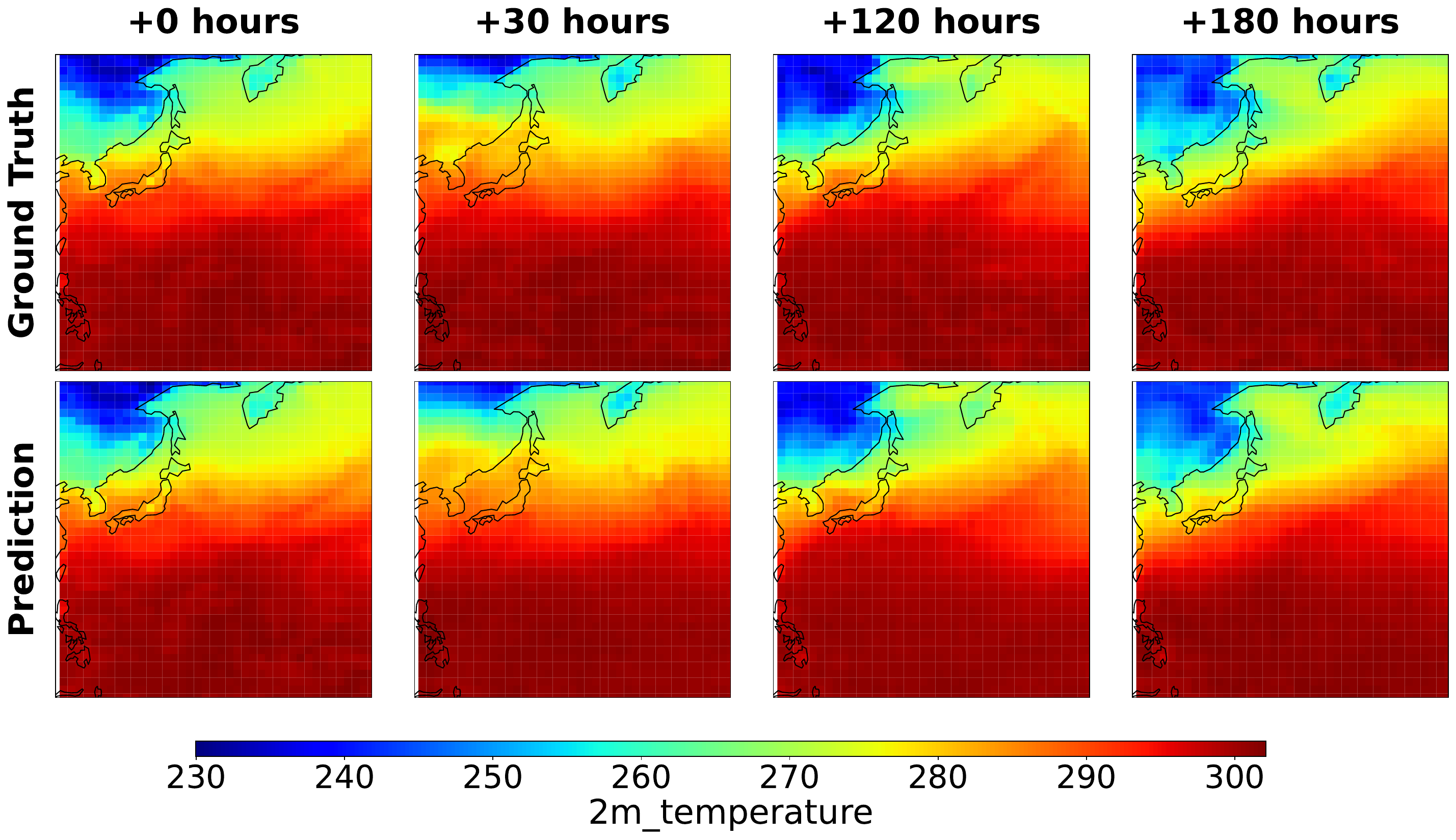}%
        \label{fig:opinf_exp1_T}}
    % \hfill
    \subfigure[Pressure prediction ($d=15$)]{
        \includegraphics[width=0.4\textwidth,height=\textheight,keepaspectratio]{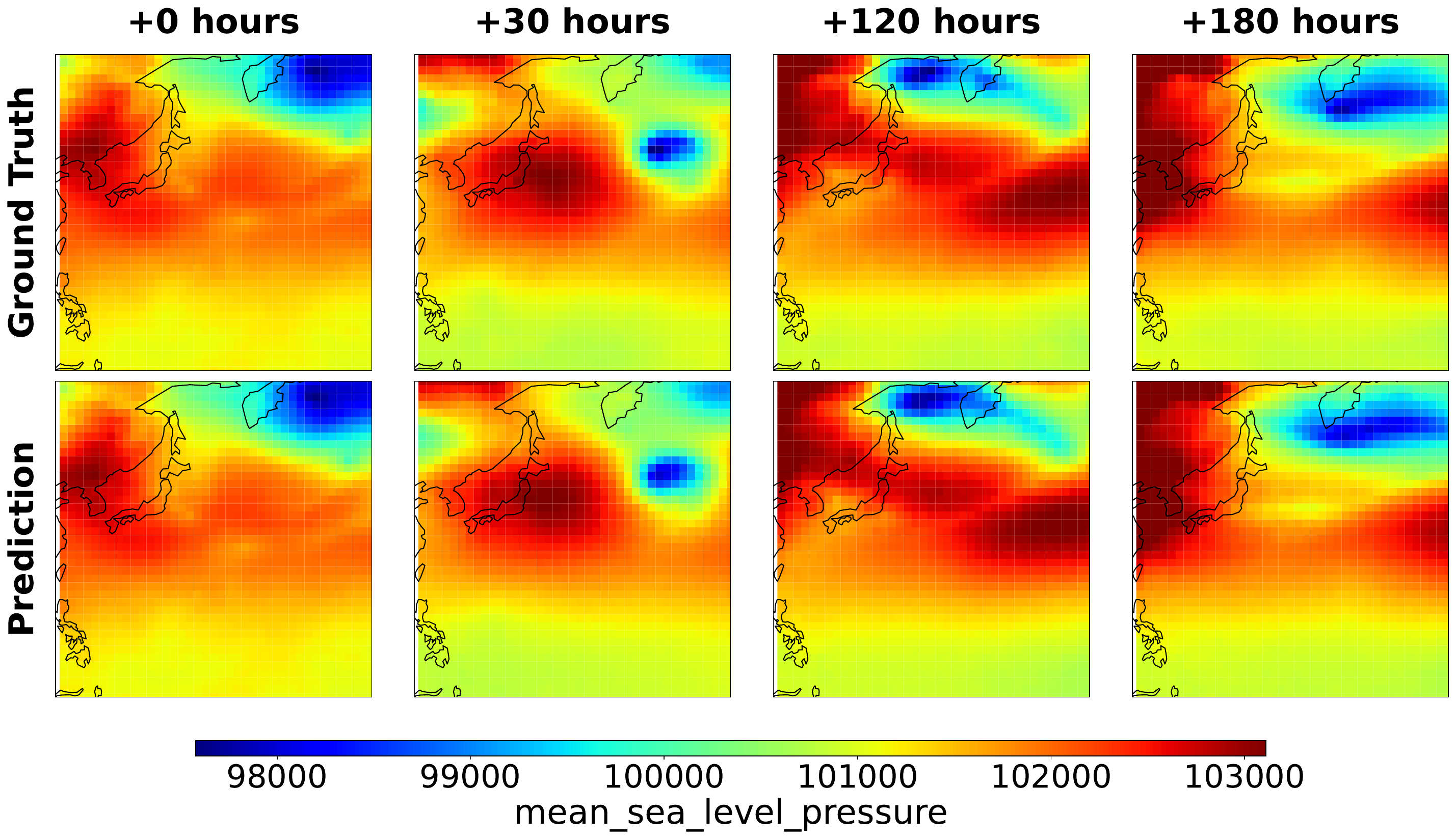}%
        \label{fig:opinf_exp1_P}}
    % \hfill
    \subfigure[u-Velocity RMSE]{
        \includegraphics[width=0.23\textwidth,height=\textheight,keepaspectratio]{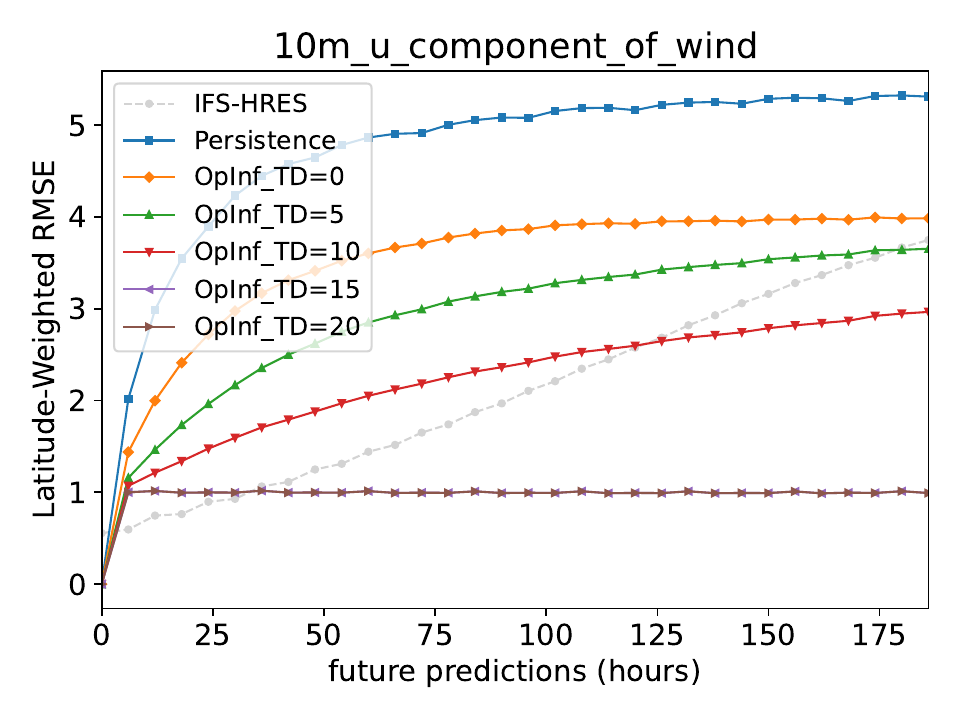}%
        \label{fig:opinf_exp1_u}}
    % \hfill
    \subfigure[v-Velocity RMSE]{
        \includegraphics[width=0.23\textwidth,height=\textheight,keepaspectratio]{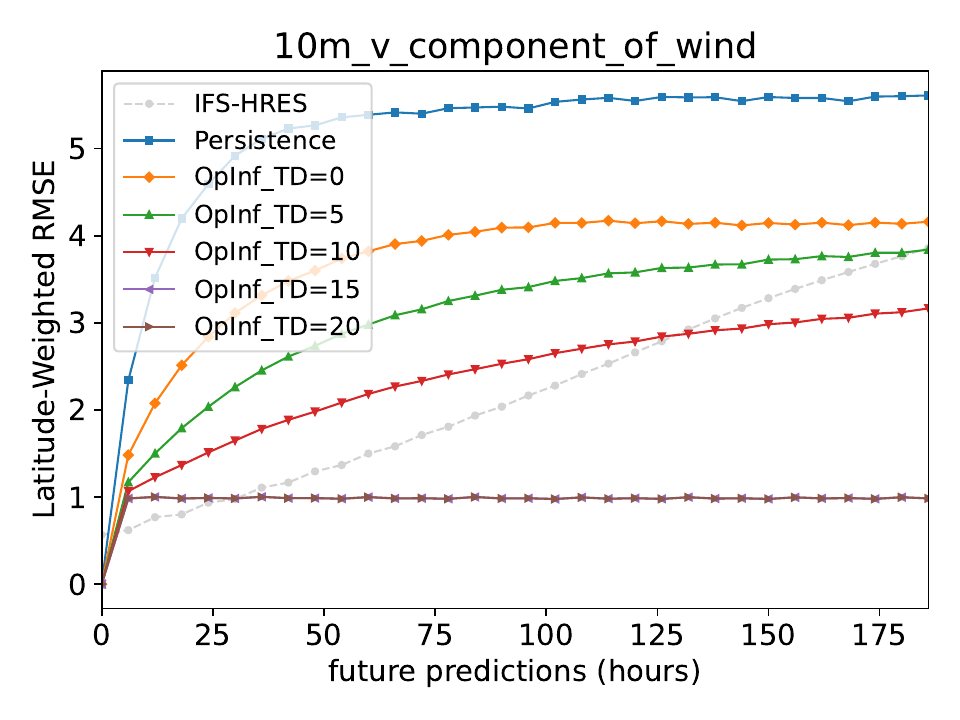}%
        \label{fig:opinf_exp1_v}}
    % \hfill
    \subfigure[Temp. RMSE]{
        \includegraphics[width=0.23\textwidth,height=\textheight,keepaspectratio]{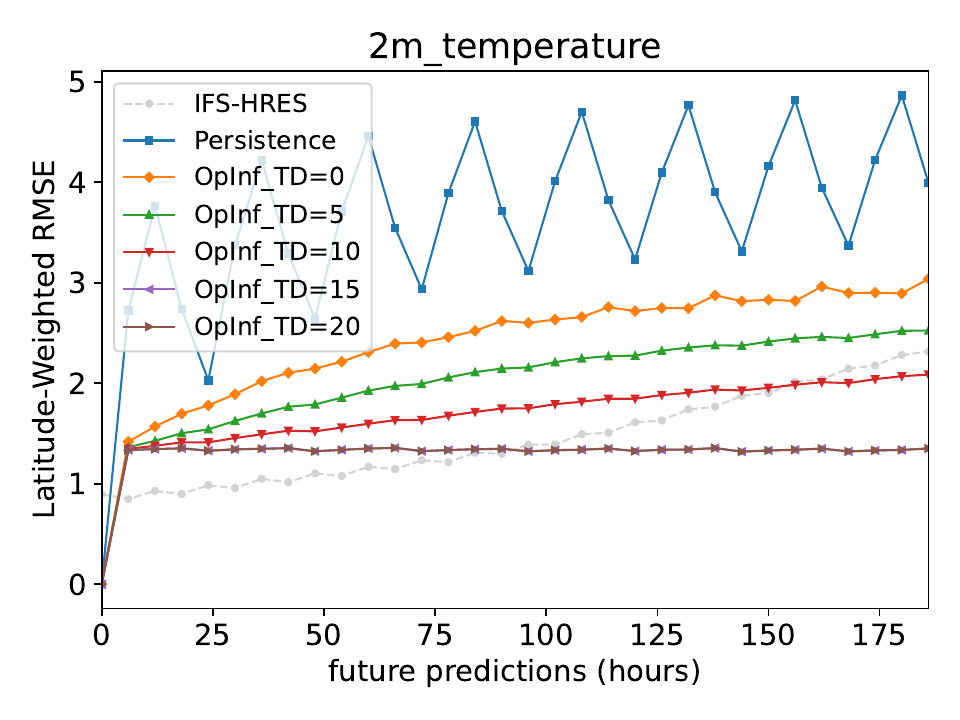}%
        \label{fig:opinf_exp1_T}}
    % \hfill
    \subfigure[Pressure RMSE]{
        \includegraphics[width=0.23\textwidth,height=\textheight,keepaspectratio]{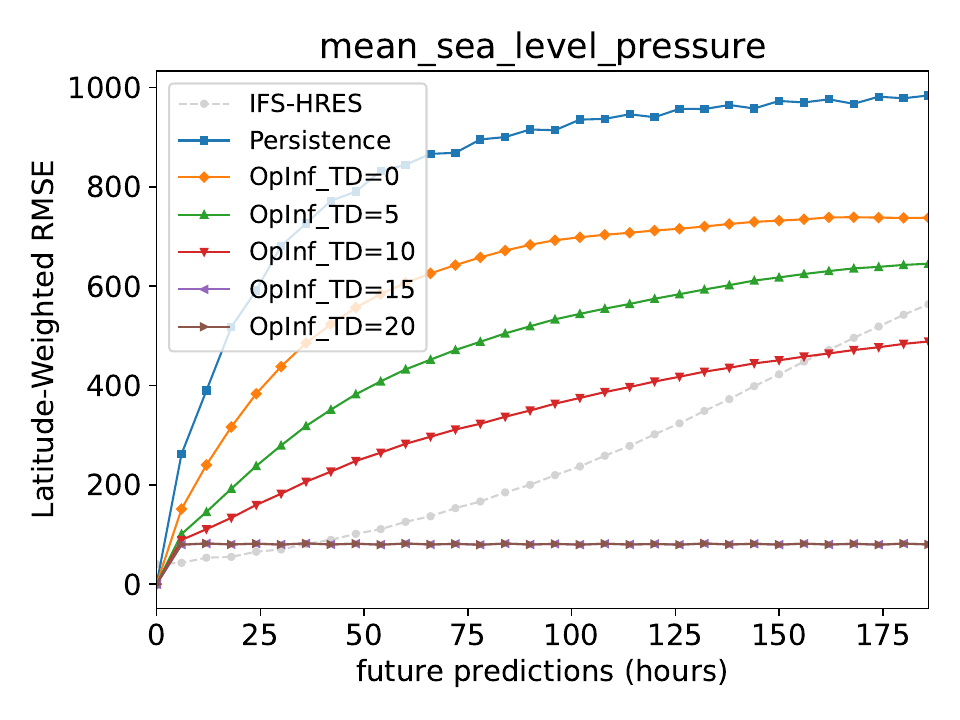}%
        \label{fig:opinf_exp1_P}}
    \vspace{-0.25cm}
    \caption{Model performance when the prediction trajectory lies within the training window.}
    \label{fig:opinf_exp1}
\end{figure}

\noindent {\bf Experiment 1: Interpolation within Training Data.} To evaluate the model's ability to learn and interpolate the dynamics within the training data, we select 100 initial conditions from the training period, each separated by 9 days, starting from December 1, 2018 (12:00AM). For each initial condition, we predict the subsequent 8 days, ensuring that all prediction trajectories remain within the training window. Figure \ref{fig:opinf_exp1} shows a single prediction trajectory and the average LW-RMSE over the forecast horizon for different numbers of time-delay steps $d$. We observe that increasing the number of time-delay steps improves the prediction accuracy within the training window. With $d=15$, the model captures the dynamics almost exactly, and the prediction error corresponds primarily to the reconstruction error from the CAE encoding.
The improvement with increasing $d$ is attributed to the model's enhanced ability to capture the system's dynamics by incorporating more historical information. Notably, with $d=15$, the number of unknowns (elements in the delayed operator) in the least-squares problem approaches the number of equations (available ERA5 snapshots), leading to an optimal fit within the training data.

\begin{figure}[ht]
    \centering
    \subfigure[u-Velocity prediction ($d=25$)]{
        \includegraphics[width=0.4\textwidth,height=\textheight,keepaspectratio]{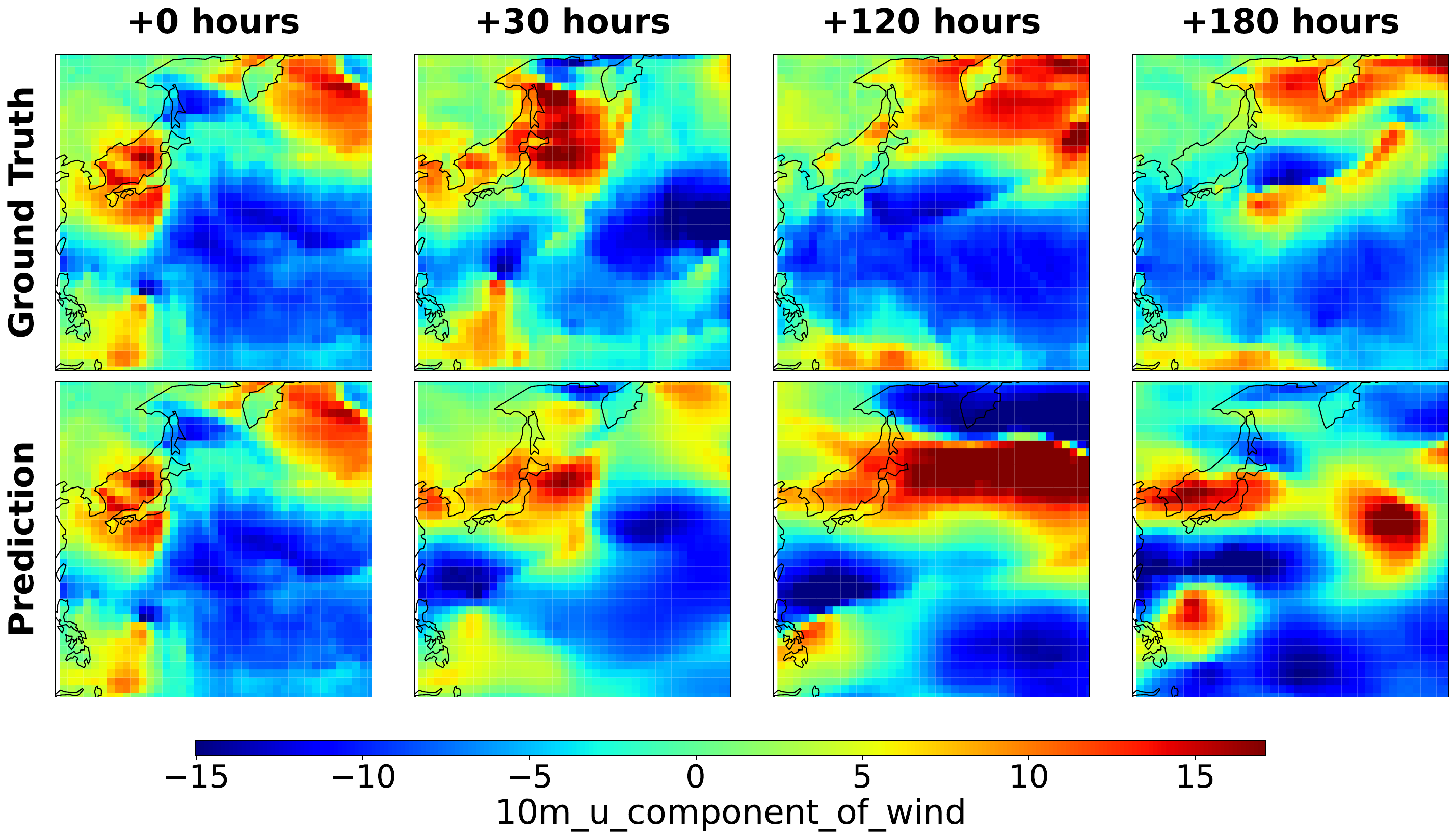}%
        \label{fig:opinf_exp2_u}}
    % \hfill
    \subfigure[v-Velocity prediction ($d=25$)]{
        \includegraphics[width=0.4\textwidth,height=\textheight,keepaspectratio]{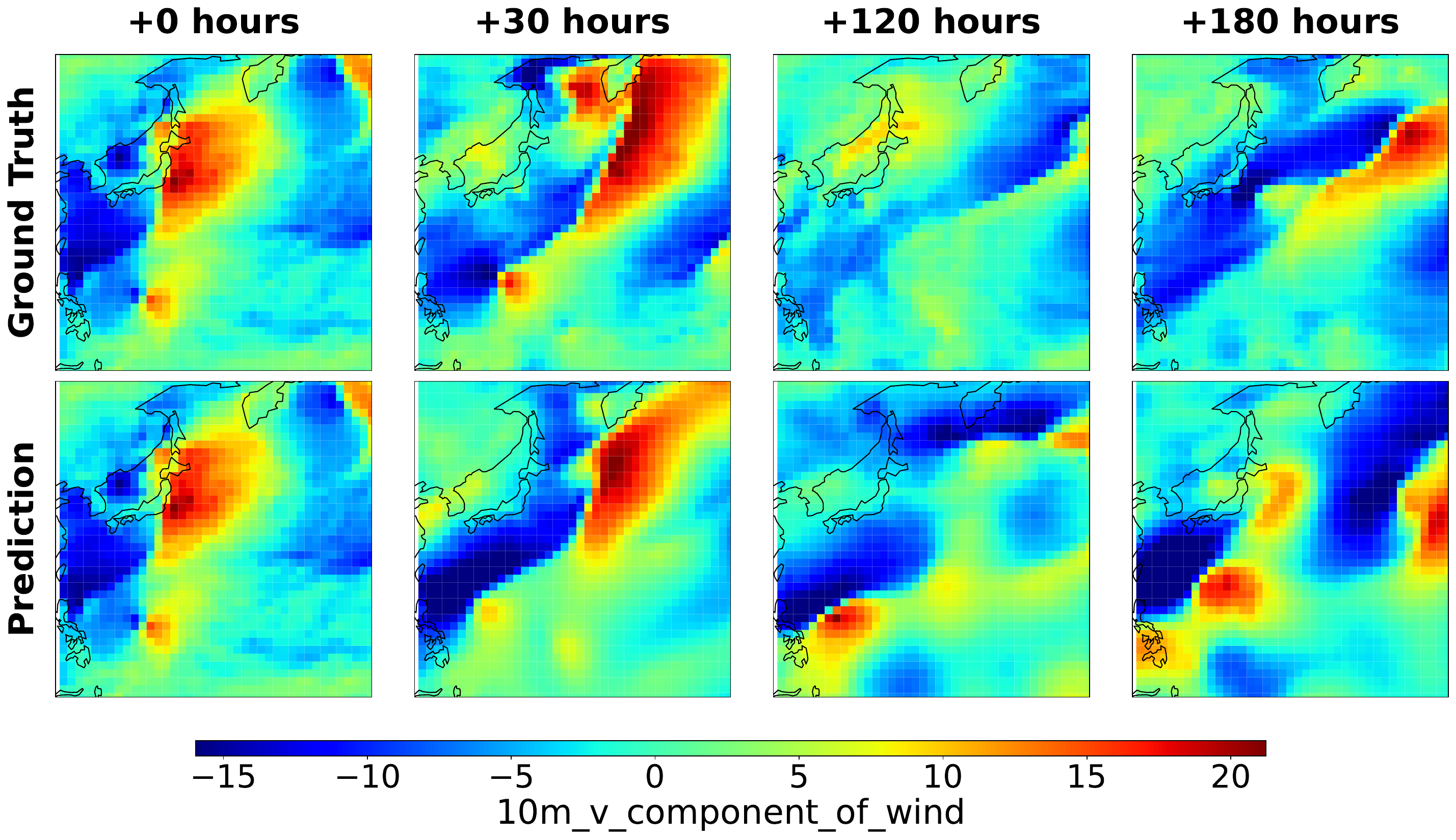}%
        \label{fig:opinf_exp2_v}}
    % \hfill
    \subfigure[Temperature prediction ($d=25$)]{
        \includegraphics[width=0.4\textwidth,height=\textheight,keepaspectratio]{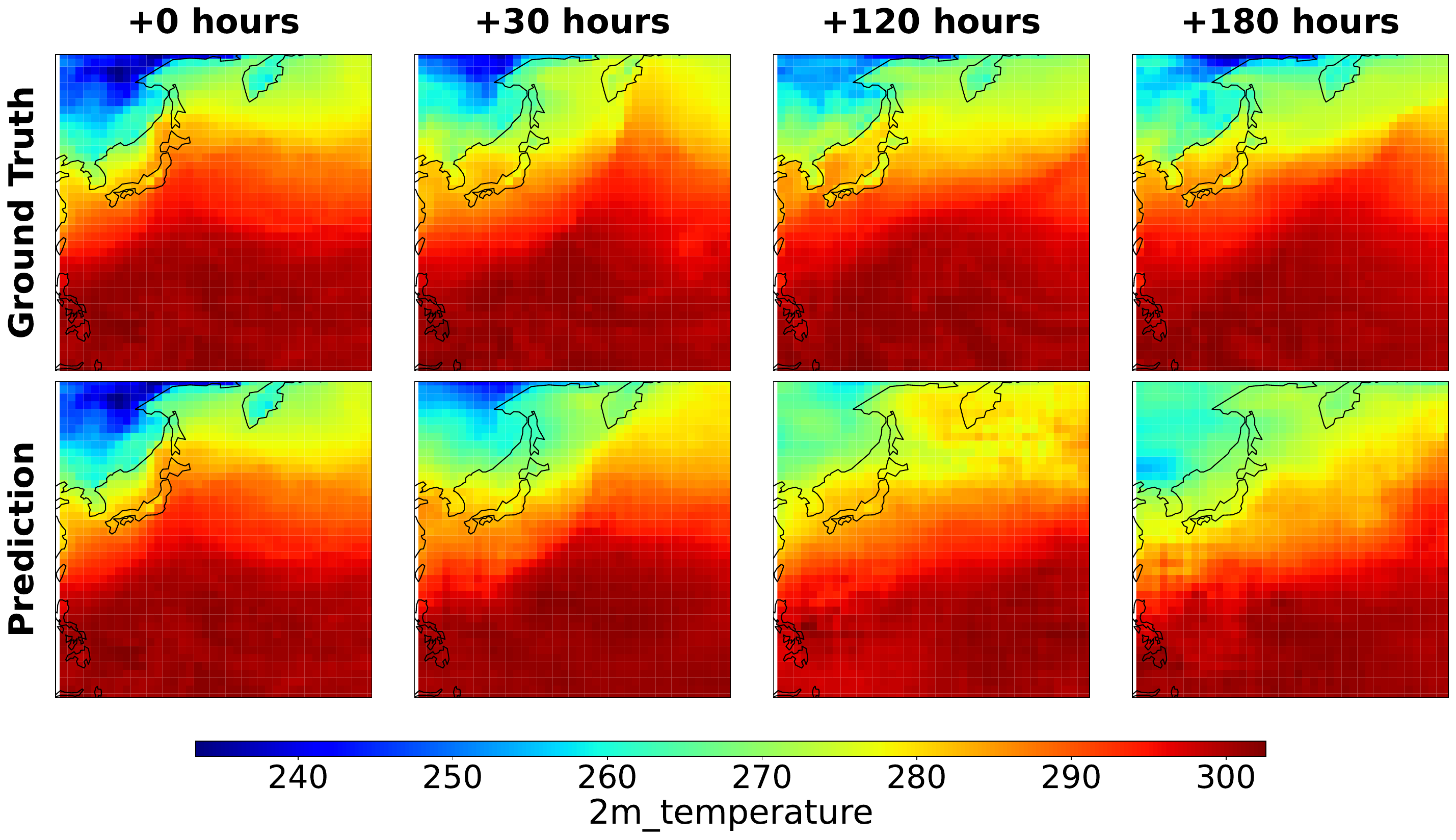}%
        \label{fig:opinf_exp2_T}}
    % \hfill
    \subfigure[Pressure prediction ($d=25$)]{
        \includegraphics[width=0.4\textwidth,height=\textheight,keepaspectratio]{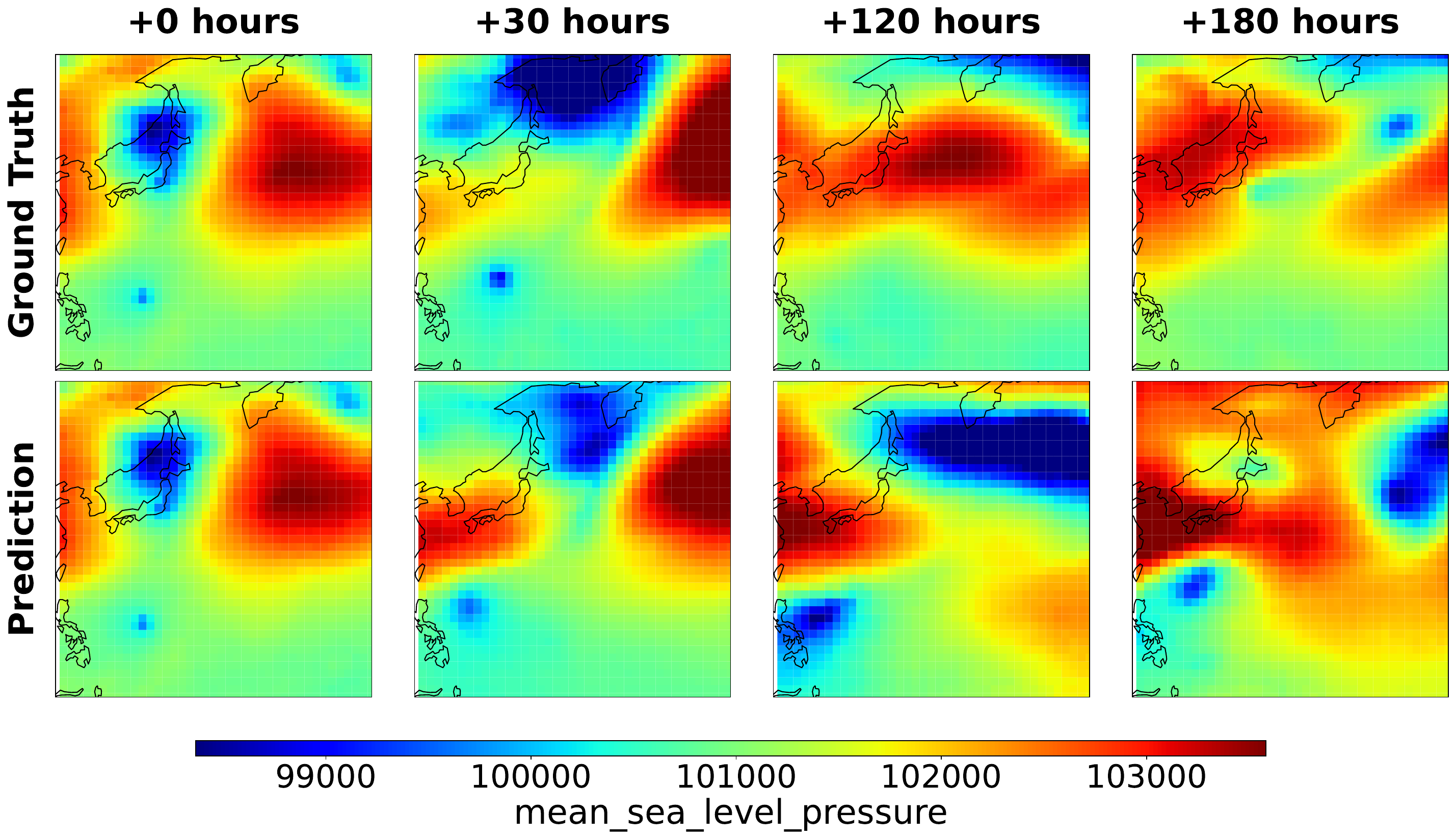}%
        \label{fig:opinf_exp2_P}}
    % \hfill
    \subfigure[u-Velocity RMSE]{
        \includegraphics[width=0.23\textwidth,height=\textheight,keepaspectratio]{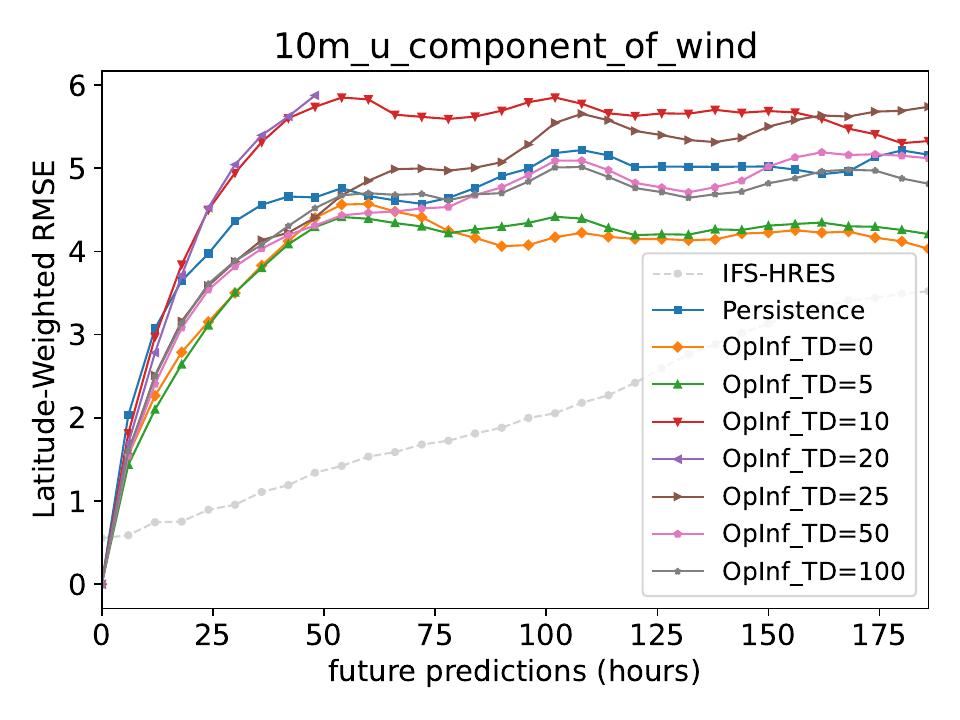}%
        \label{fig:opinf_exp2_u}}
    % \hfill
    \subfigure[v-Velocity RMSE]{
        \includegraphics[width=0.23\textwidth,height=\textheight,keepaspectratio]{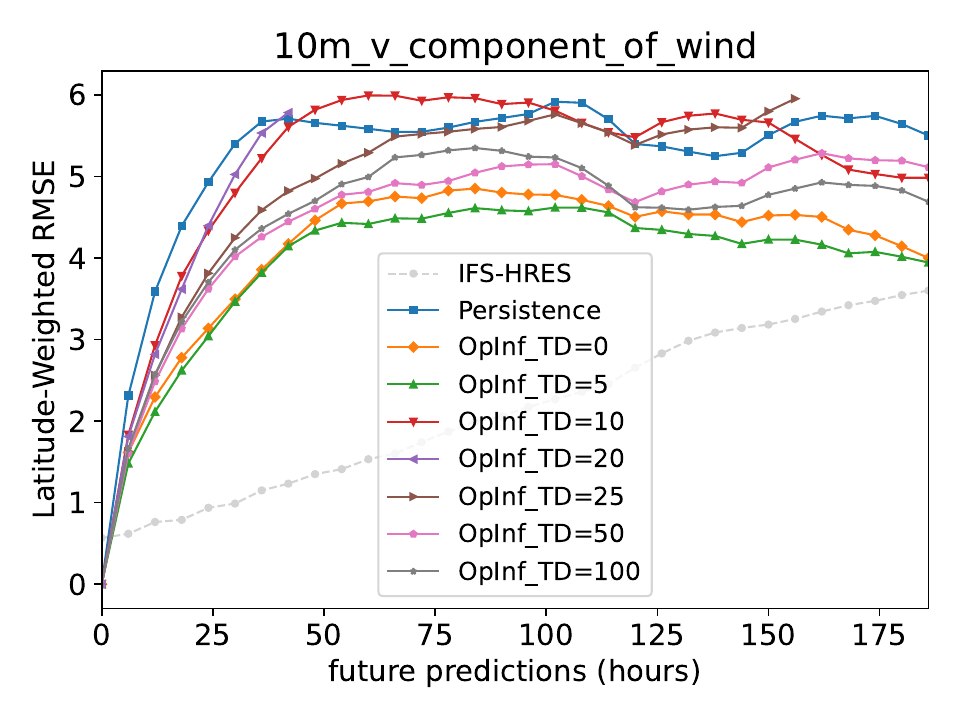}%
        \label{fig:opinf_exp2_v}}
    % \hfill
    \subfigure[Temp. RMSE]{
        \includegraphics[width=0.23\textwidth,height=\textheight,keepaspectratio]{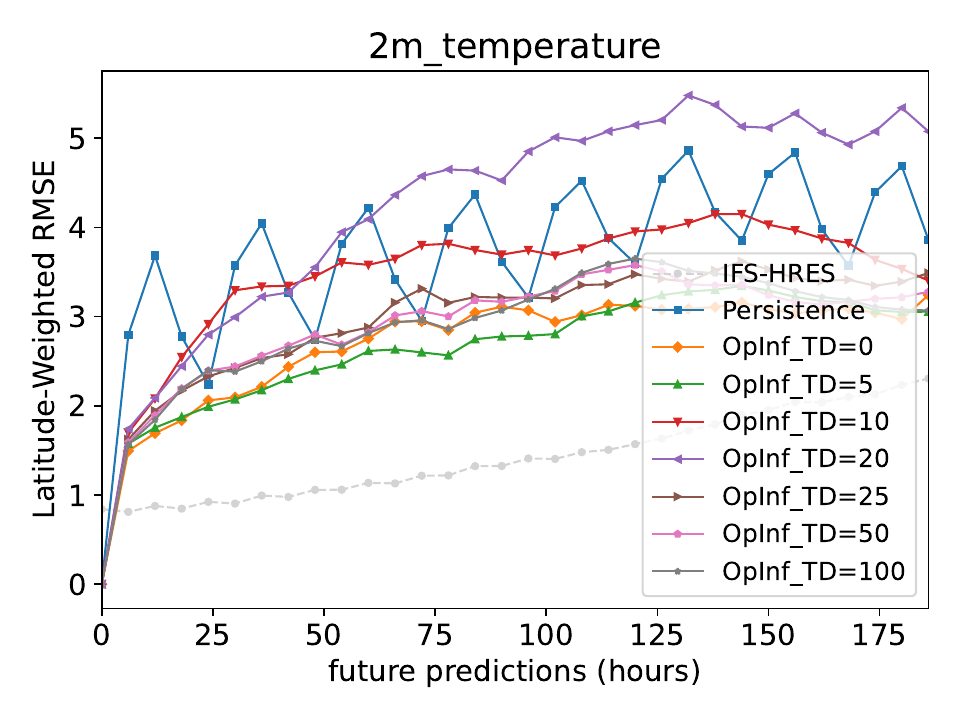}%
        \label{fig:opinf_exp2_T}}
    % \hfill
    \subfigure[Pressure RMSE]{
        \includegraphics[width=0.23\textwidth,height=\textheight,keepaspectratio]{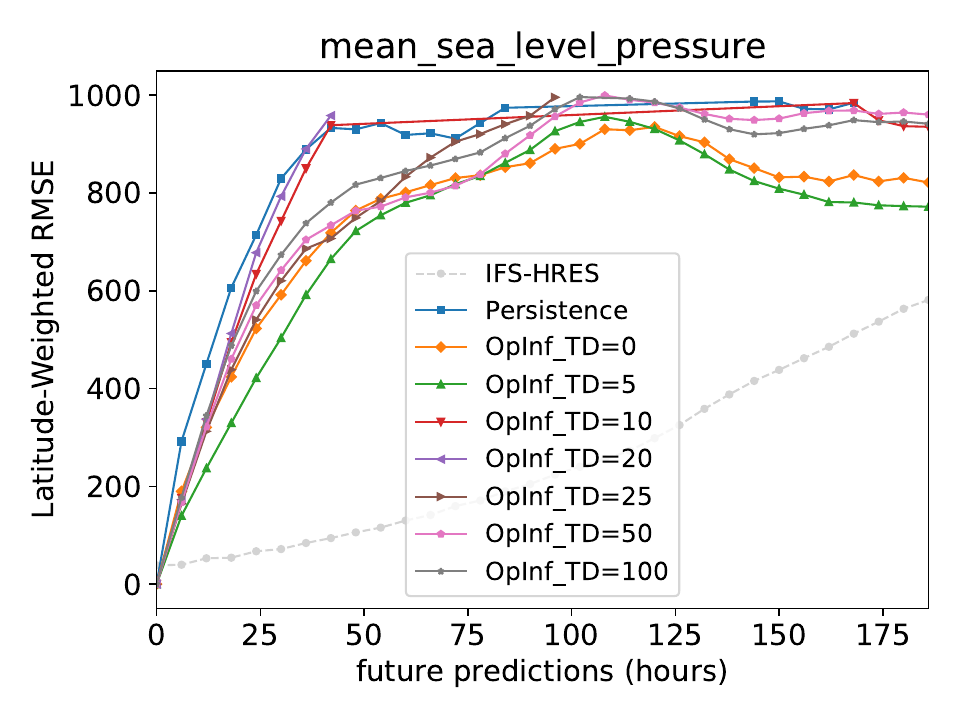}%
        \label{fig:opinf_exp2_P}}
    \vspace{-0.25cm}
    \caption{Model performance when the prediction trajectory lies outside the training window.}
    \label{fig:opinf_exp2}
\end{figure}

\noindent {\bf Experiment 2: Generalization to Unseen Future States.}
In this experiment, we assess the model's ability to generalize to unseen future states by starting with an initial condition outside the training window and predicting the next 8 days. Figure \ref{fig:opinf_exp2} illustrates the predicted fields and the LW-RMSE for each variable over the forecast horizon.
%The results show a significant jump in RMSE at the beginning of the forecast, with errors increasing over time. This indicates that while the model has effectively learned to interpolate within the training data, it struggles to generalize to future states not seen during training. The performance, although better than a persistence model (which predicts no change from the initial condition), is insufficient for reliable out-of-sample predictions.
The  evaluation reveals two key limitations in the model's predictive capabilities. First, the immediate spike in RMSE at the initiation of the forecast suggests fundamental challenges in the model's ability to project beyond its training data. Second, the steady deterioration in accuracy over the forecast period indicates persistent issues with long-term prediction reliability. While the model demonstrates superior performance compared to baseline persistence forecasts, its current predictive accuracy falls short of operational requirements.

\begin{figure}[ht]
    \centering
    \subfigure[u-Velocity prediction ($d=5$)]{
        \includegraphics[width=0.4\textwidth,height=\textheight,keepaspectratio]{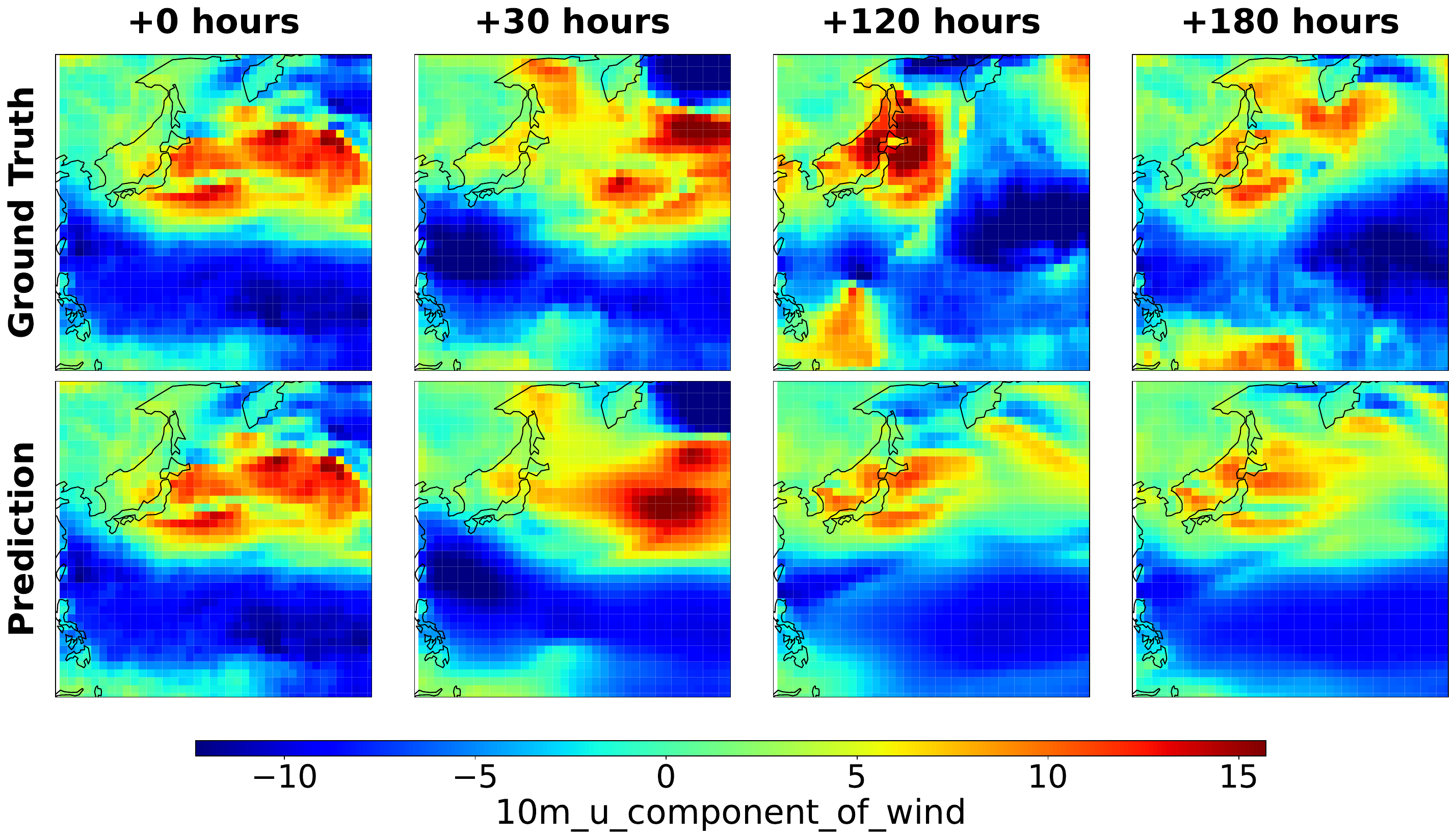}%
        \label{fig:opinf_exp3_u}}
    % \hfill
    \subfigure[v-Velocity prediction ($d=5$)]{
        \includegraphics[width=0.4\textwidth,height=\textheight,keepaspectratio]{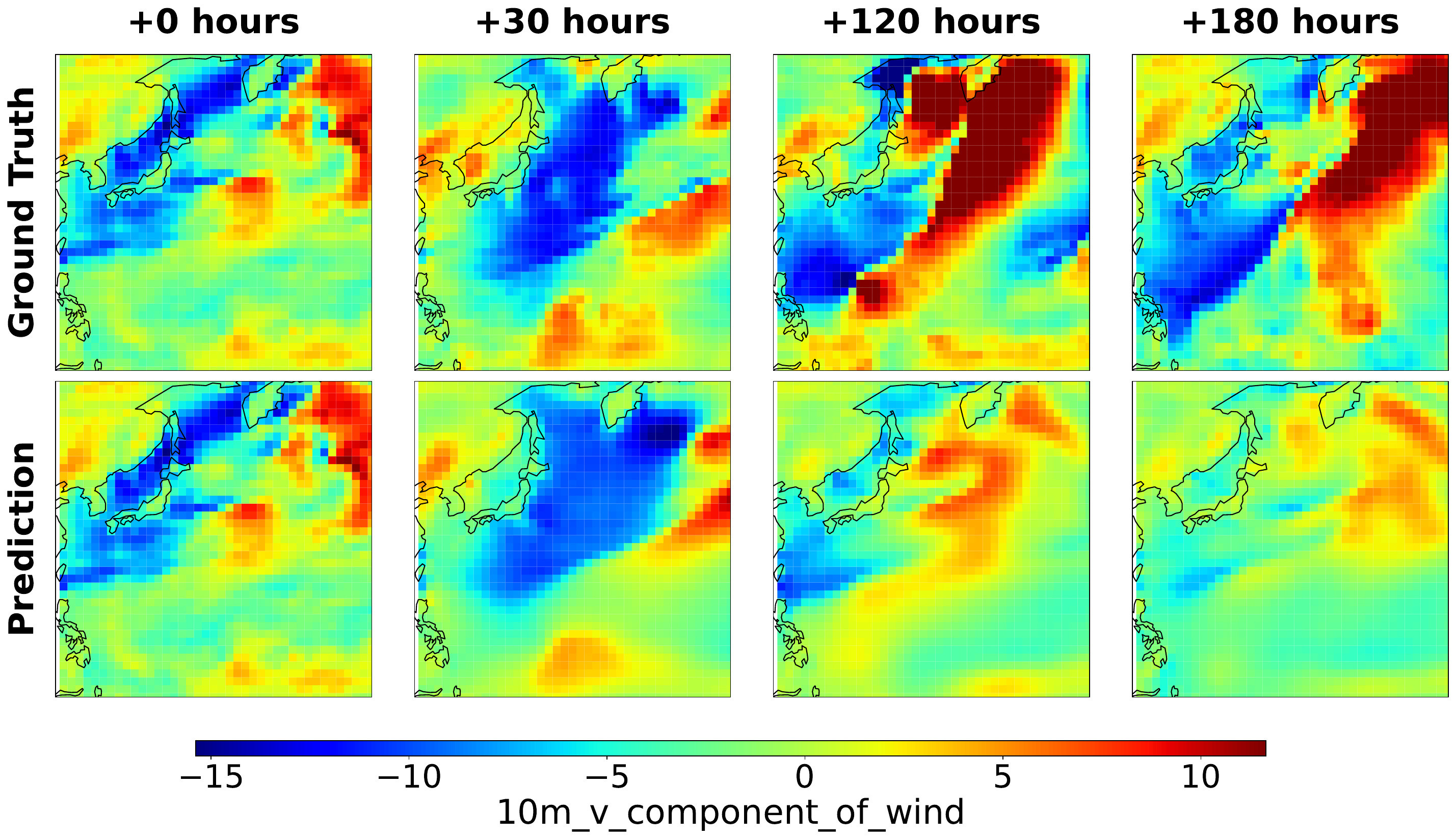}%
        \label{fig:opinf_exp3_v}}
    % \hfill
    \subfigure[Temperature prediction ($d=5$)]{
        \includegraphics[width=0.4\textwidth,height=\textheight,keepaspectratio]{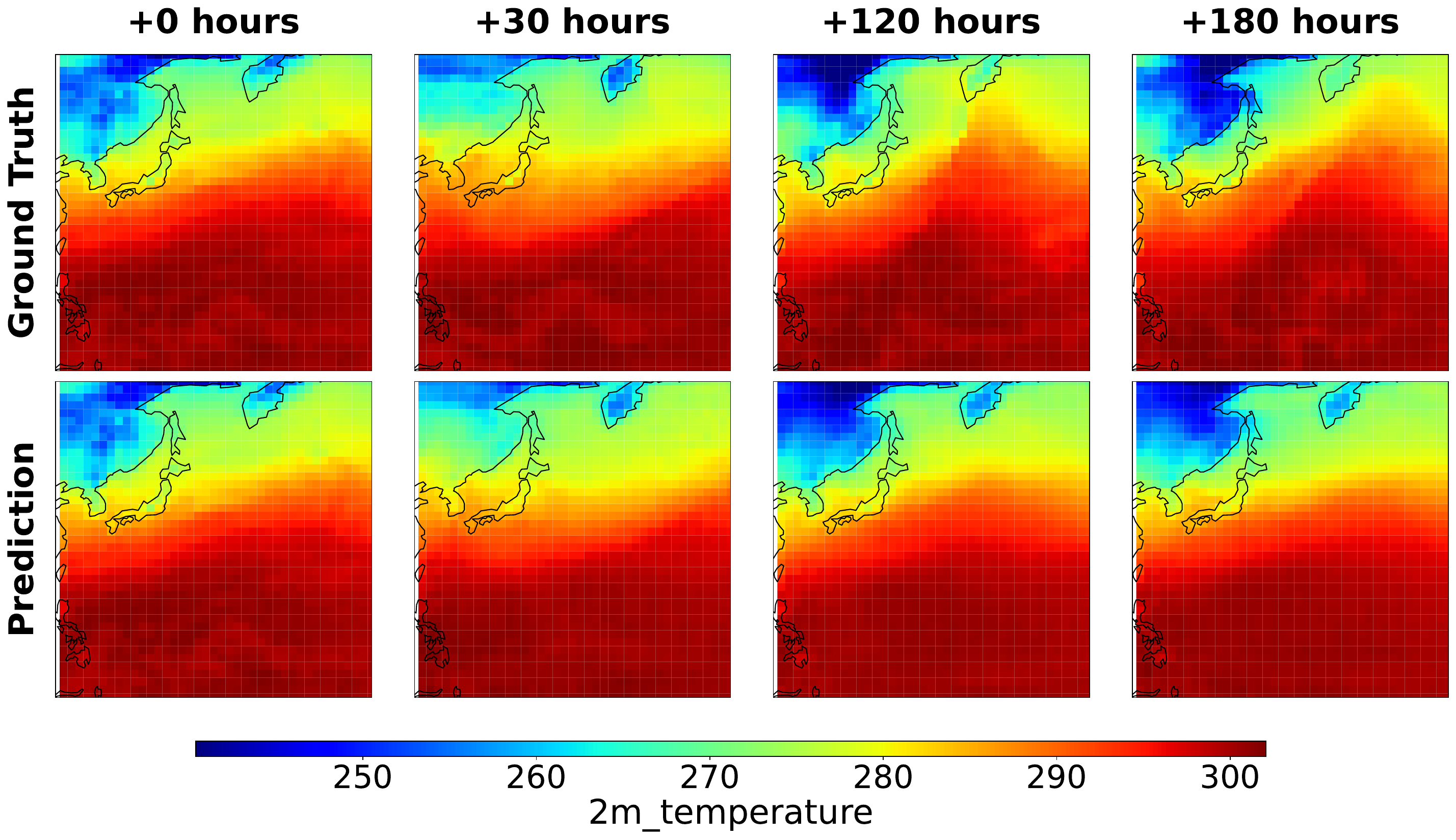}%
        \label{fig:opinf_exp3_T}}
    % \hfill
    \subfigure[Pressure prediction ($d=5$)]{
        \includegraphics[width=0.4\textwidth,height=\textheight,keepaspectratio]{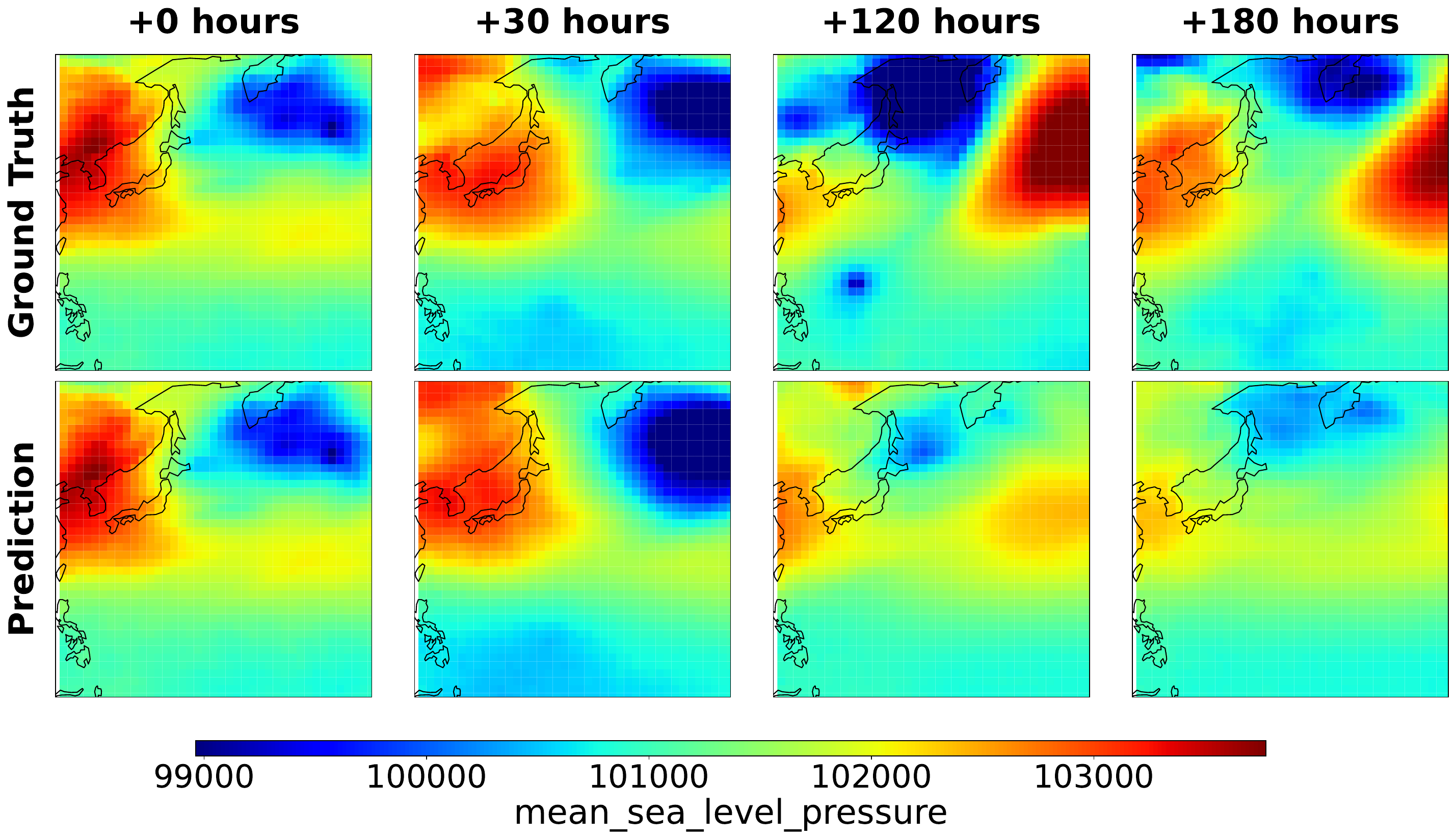}%
        \label{fig:opinf_exp3_P}}
    % \hfill
    \subfigure[u-Velocity RMSE]{
        \includegraphics[width=0.23\textwidth,height=\textheight,keepaspectratio]{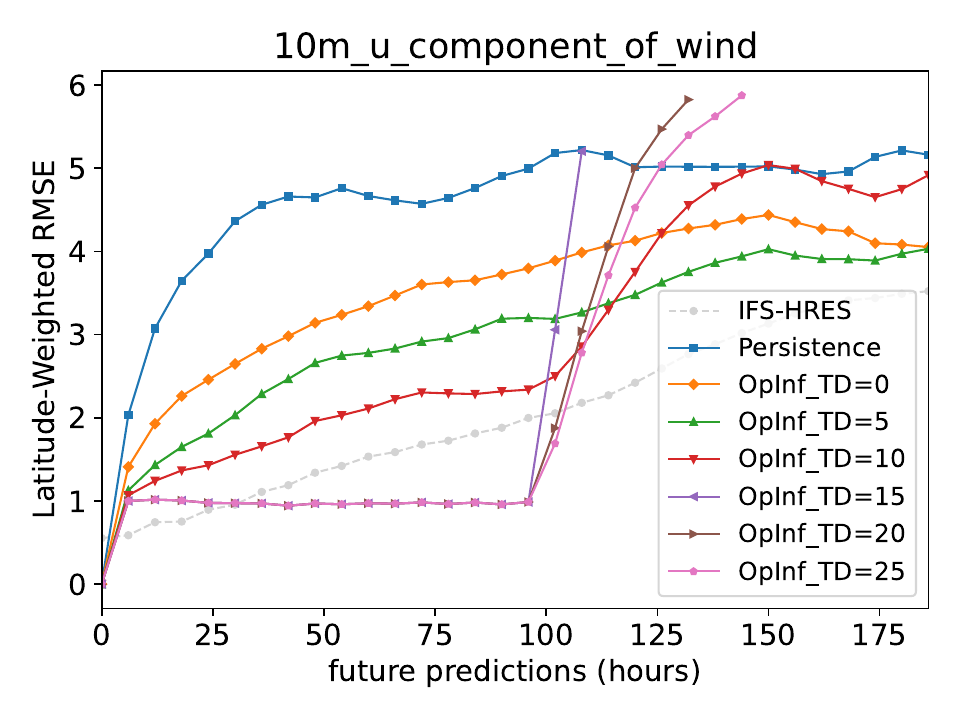}%
        \label{fig:opinf_exp3_u}}
    % \hfill
    \subfigure[v-Velocity RMSE]{
        \includegraphics[width=0.23\textwidth,height=\textheight,keepaspectratio]{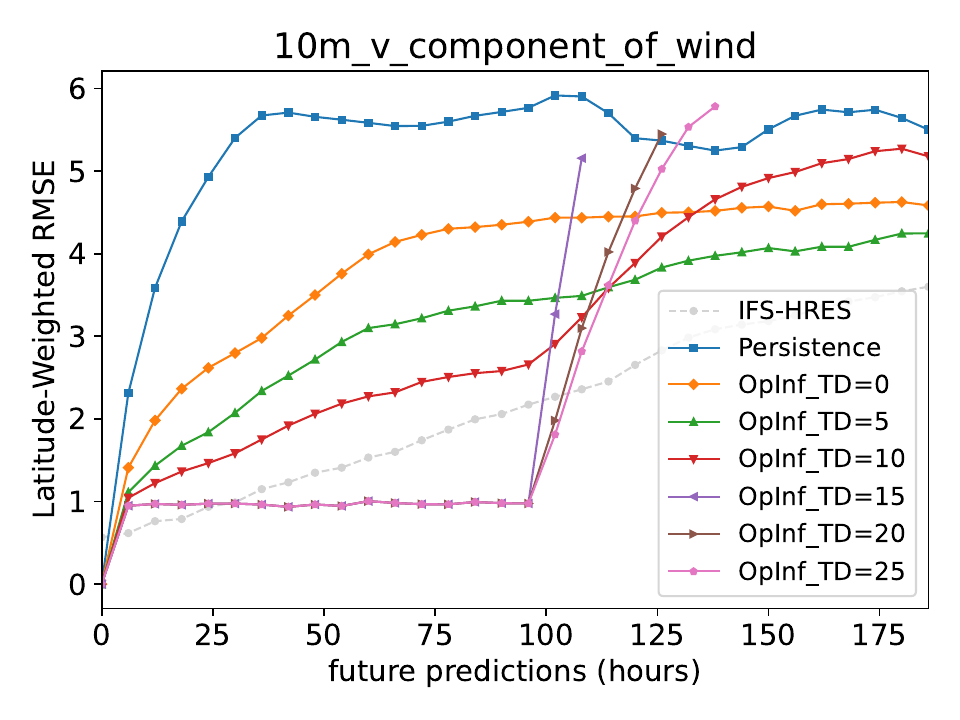}%
        \label{fig:opinf_exp3_v}}
    % \hfill
    \subfigure[Temp. RMSE]{
        \includegraphics[width=0.23\textwidth,height=\textheight,keepaspectratio]{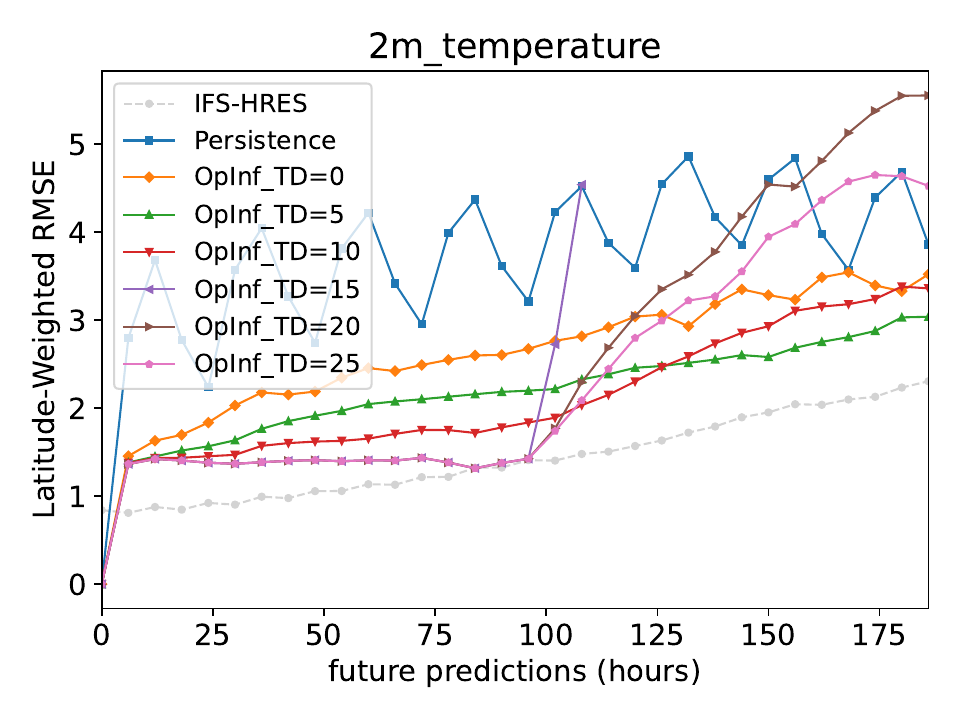}%
        \label{fig:opinf_exp3_T}}
    % \hfill
    \subfigure[Pressure RMSE]{
        \includegraphics[width=0.23\textwidth,height=\textheight,keepaspectratio]{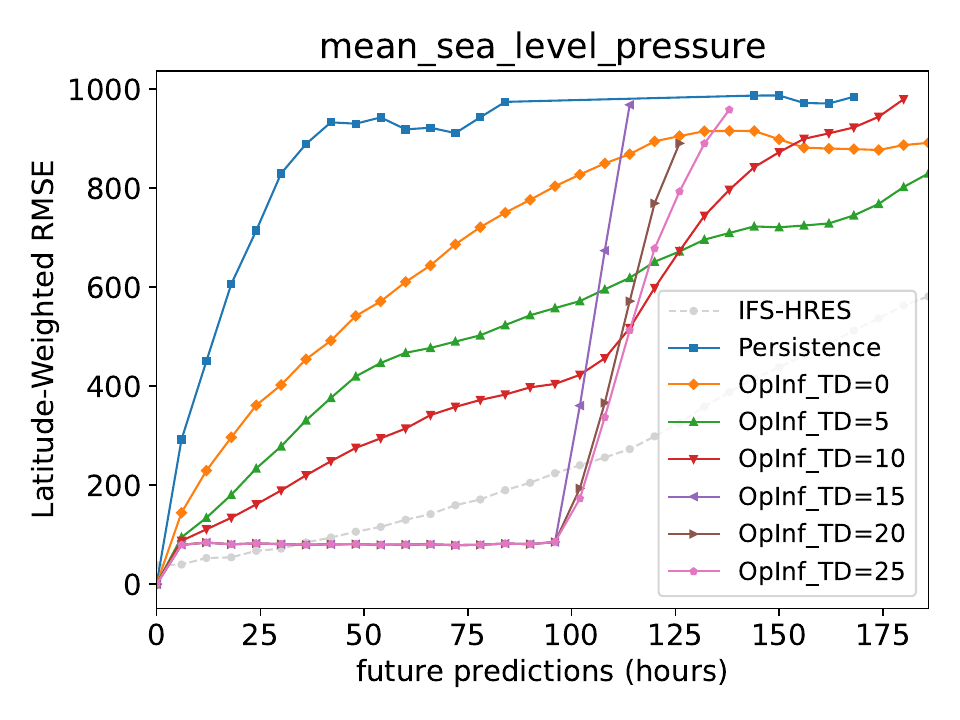}%
        \label{fig:opinf_exp3_P}}
    \vspace{-0.25cm}
    \caption{Model performance when the prediction trajectory starts within the training window and transitions to unseen future states.}
    \label{fig:opinf_exp3}
    \vspace{-0.5cm}
\end{figure}

\noindent {\bf Experiment 3: Transition from Training to Unseen States.}
To further investigate the model's generalization capabilities, we start with an initial condition near the end of the training window, such that the first half of the 8-day prediction lies within the training window, and the second half extends into the unseen future. Figure \ref{fig:opinf_exp3} shows the predicted states and the LW-RMSE over the forecast trajectory for this setup. As expected, the model maintains low prediction errors while the forecast remains within the training window. However, once the prediction enters the unseen future states, the RMSE increases sharply, mirroring the behavior observed in Experiment 2. This further confirms the model's limitations in generalizing beyond the training data.

\vspace{-0.5cm}
\section{Conclusion}
\label{sec:conclusion}
In this study, we presented an efficient reduced-order modeling (ROM) framework for short-range weather prediction by integrating Convolutional Autoencoders (CAEs) enhanced with Convolutional Block Attention Modules (CBAM) and operator inference augmented with time-delay embeddings. We highlight the key takeaways from our experiments:
\begin{itemize}
\setlength\itemsep{-0.2cm}
    \item Reconstruction quality, rather than inference methodology, represents the primary bottleneck for prediction accuracy. The model's ability to accurately project future atmospheric states from its compressed representation fundamentally limits its predictive capabilities. This finding suggests that future research could focus on developing more sophisticated dimensionality reduction techniques rather than refining neural time-stepping mechanisms.
    \item Importantly, we found that traditional error metrics such as RMSE do not effectively capture the quality of weather predictions. Even when RMSE values  appear reasonable, they often fail to reflect important aspects of the prediction, such as the preservation of coherent atmospheric structures or the capture of extreme events. This indicates a need for more sophisticated evaluation metrics that better align with meteorological significance.
    \item The relationship between time-delay embeddings and prediction accuracy reveals something fundamental about weather systems. When predicting within the training window, increasing the number of time-delay steps dramatically improves accuracy. This suggests that weather patterns contain strong temporal correlations that can be captured through linear operations in the right embedding space.
    \item The dimensionality reduction analysis shows that even with attention mechanisms, compressing the atmospheric state into a manageable latent space (960 dimensions) loses crucial information needed for long-term prediction, thus motivating the need for richer and more sophisticated dimensionality-reduction formulations with structured latent spaces.
\end{itemize}
\vspace{-0.2cm}

With the context that these ROMs are easily trained in a few hours on a single GPU (in the case of POD, a few seconds), these insights suggest a promising direction: ROM frameworks can serve as computationally efficient engines for capturing dominant  weather (and potentially climate) patterns, while state-of-the-art AI models can act as targeted corrective layers that resolve finer-scale atmospheric phenomena missed by dimensional reduction. We recognize that significant research is needed to effectively bridge these methodologies.

\clearpage
\acks{}
The authors acknowledge support from the AFOSR grant FA9550-17-1-0195 and from Los Alamos National Laboratory under the grant `Algorithm/Software/Hardware Co-design for High-Energy Density Applications.'

\bibliography{ref}

\end{document}